\documentclass[preprint,3p,authoryear,times]{elsarticle}

\journal{} 
\usepackage[fleqn]{amsmath}
\usepackage[]{amsfonts}
\usepackage[]{amsthm}
\usepackage[]{amssymb}
\usepackage{empheq}
\usepackage{lscape} 

\usepackage[mathlines]{lineno}

\usepackage[]{algorithm2e}
\usepackage[english]{babel}
\usepackage[T1]{fontenc}

\usepackage{hyperref} 
\usepackage{graphicx}
\usepackage{caption,subcaption}
\usepackage{tikz}
\usepackage{pstricks}
\usepackage{xstring}
\usepackage{multirow}
\usepackage{hhline}
\usepackage{natbib} 

\definecolor{darkblue}{rgb}{0,0,0.5}

\hypersetup{colorlinks=true,linkcolor=black,citecolor=darkblue,urlcolor=darkblue} 

\newcommand{\newref}[2]{\hyperref[#2]{#1~\ref*{#2}}} 

\usepackage{longtable}
\usepackage{pdflscape}
\usepackage{tabularx}
\usepackage[toc,page]{appendix}

\usepackage{float}

\begin{document}
\begin{frontmatter}

\title{Dynamic multi-agent deep reinforcement learning-based pricing and incentivization approach in multimodal transportation networks}

\author[1]{Khadidja Kadem\corref{cor}}
\cortext[cor]{Corresponding author.} 
\ead{khadidja.kadem@univ-eiffel.fr}
\author[1,2]{Mostafa Ameli}
\author[3]{Carlos Lima Azevedo}
\author[1]{Mahdi Zargayouna} 
\author[1]{Latifa Oukhellou}

\address[1]{COSYS-GRETTIA, Univ Gustave Eiffel, Marne-la-Vallée, France}
\address[2]{Department of Electrical Engineering and Computer Sciences,  University of California, Berkeley, Berkeley, USA} 
\address[3]{Department of Technology, Management and Economics, Technical University of Denmark, Denmark} 

\begin{sloppypar} 
\begin{abstract} 
In multimodal transportation systems, shared mobility services (SMSs) are often promoted for their potential to enhance flexibility and reduce congestion. However, SMSs demand is often concentrated in high-density areas, which can limit the effectiveness and accessibility for various commuter groups across a city. This uneven integration challenges the efficiency of the transportation system, especially in terms of emissions and spatial equity. Addressing these issues requires coordination among multiple stakeholders whose objectives frequently conflict. Whereas authorities aim to ensure sustainable and equitable mobility, SMSs providers focus on revenue maximization, and travelers seek to minimize personal travel costs. This paper proposes a multi-agent deep reinforcement learning framework that captures these interactions and reconciles competing goals through dynamic pricing and incentivization strategies for SMSs and public transport. The framework integrates multimodal macroscopic simulation with two reinforcement learning (RL) agents: (i) a public authority that allocates spatio-temporal public transport incentives to improve equity, emissions, and efficiency, and (ii) an SMS provider that dynamically adjusts its fares to optimize revenue. 
The agents interact iteratively with the simulated transportation system to learn optimal strategies in response to evolving demand, congestion, and network conditions. Numerical experiments conducted over a three-hour morning peak period, at a 20-minute temporal resolution, show that dynamic incentivization effectively reduces congestion peaks, lowers commuters' costs by around 20 \% and emissions by approximately 10 \%, while nearly doubling public transport profit and supporting a more equitable distribution of benefits. When combined with dynamic SMS pricing, the two RL agents demonstrate the capacity to balance conflicting objectives between private providers and public authorities. The proposed approach provides a decision-support tool for guiding sustainable and equitable multimodal mobility planning.

\end{abstract}


\end{sloppypar} 
\end{frontmatter}


\begin{sloppypar} 

\section{Introduction}

The rapid growth of the urban population is intensifying pressure on cities to develop more sustainable and space-efficient mobility systems. Today, private vehicles continue to dominate urban travel, consuming significantly more road space and energy than public transport, and contributing to traffic congestion and climate change \citep{rode_accessibility_2017}. In fact, road transport is responsible for nearly 30\% of greenhouse gas emissions in Europe \citep{eu_emissions_2024}. In this context, shared mobility services (SMSs), such as ridesharing and carpooling, offer a promising alternative by enabling higher vehicle occupancy, reducing the number of vehicles on the road, and complementing public transit networks. Moreover, SMSs can help address accessibility gaps by providing mobility options in areas underserved by fixed-route transit and offering more flexible and affordable travel options.

However, as recent studies suggest, the introduction of SMSs may lead to both positive and negative effects on social and environmental conditions \citep{oh_assessing_2020}. For instance, while SMSs might attract users from private vehicles, they may also compete with existing public transit. Furthermore, the demand for SMSs often concentrates in high-density areas, which can increase existing spatial and social disparities if not properly managed \citep{gao_regulating_2024}. To support informed planning and policymaking, understanding the multifaceted impacts of SMSs on the transportation system is crucial for determining how they can be effectively regulated and coordinated with public transport in ways that preserve social and environmental integrity.

While the environmental impacts of transportation are often at the forefront of policy discussions, this focus alone is insufficient to promote sustainable systems. As SMSs alter both the availability and affordability of transportation, they may reshape the way travel benefits and costs are distributed across different populations and urban areas. This raises fundamental questions: who gains the most from the availability of SMSs? In what urban areas are these benefits concentrated? And to what extent are these benefits equitably distributed? These questions are directly related to the concept of equity, as defined in the transportation literature \citep{litman_evaluating_2017}.

\begin{figure}[ht]
    \centering
    \includegraphics[width=.8\linewidth]{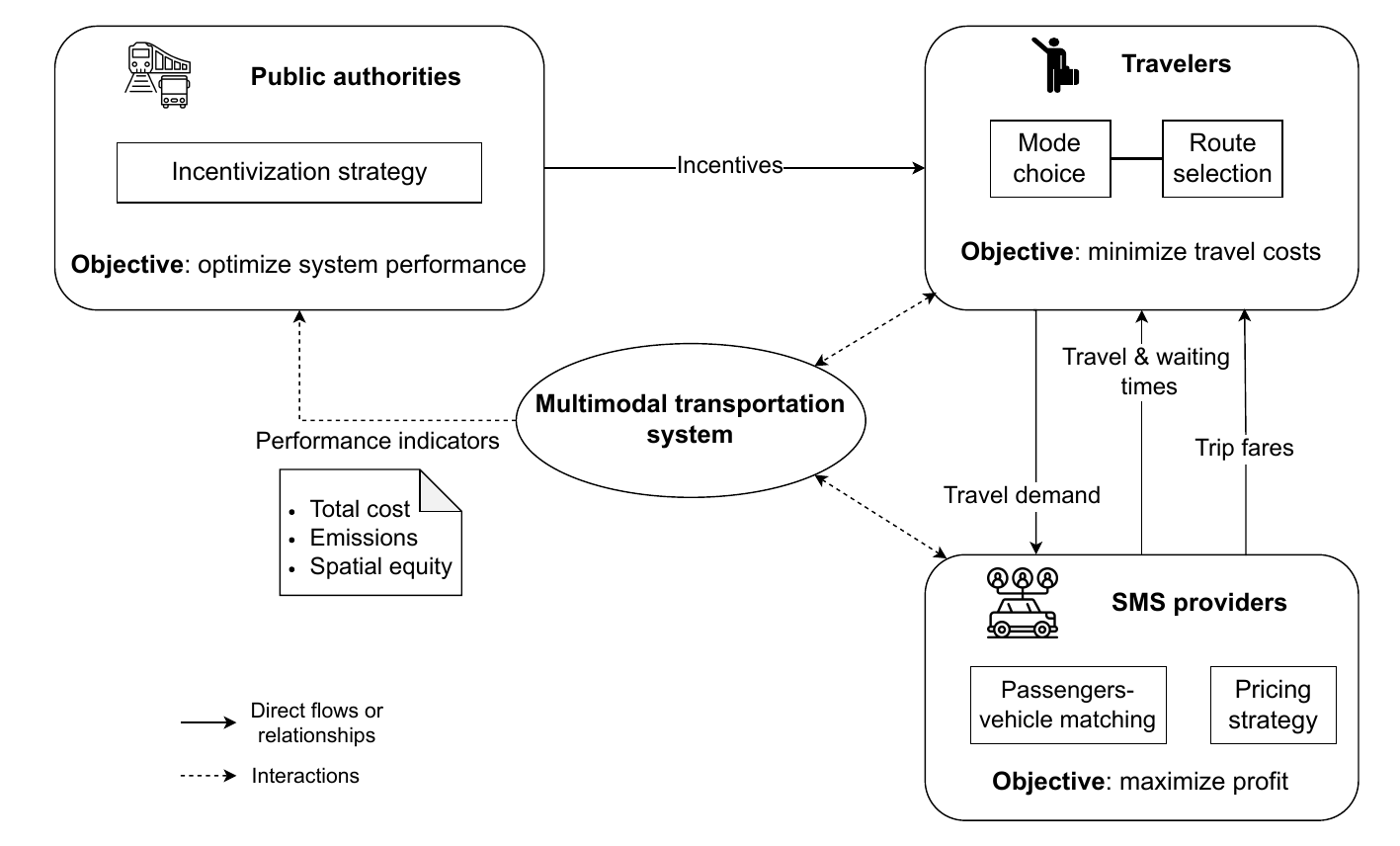}
    \caption{Stakeholder interactions and feedback mechanisms in multimodal systems with SMSs}
    \label{fig:stakeholders}
\end{figure}

Evaluating the full impacts of SMSs, therefore, requires a comprehensive modeling approach that reflects the complexity of urban systems. Such systems are characterized by multiple stakeholders, whose objectives are often misaligned, as illustrated in Figure \ref{fig:stakeholders}. On the one hand, travelers make daily travel decisions, such as mode and route selection. These decisions are influenced by factors such as income level, travel costs, and the availability of travel options. On the other hand, public authorities are responsible for managing the overall system, striving to ensure accessibility, minimize congestion, reduce emissions, and promote equity. Through incentivization policies, they can influence user behavior toward more sustainable and socially beneficial outcomes. Additionally, SMSs providers act as profit-driven agents seeking to satisfy demand and optimize their operations. These providers often employ strategies such as dynamic pricing to manage supply-demand imbalances and maximize their revenue.

Capturing the dynamics of such a complex system presents several challenges. It requires representing the interactions between various actors and travel modes, accommodating diverse interests, and representing how individual travel decisions both shape and respond to the overall performance of the transportation network. In light of these challenges, this study aims to achieve two primary objectives. The first is to propose a modeling framework that enables policymakers to rigorously assess the implications of introducing SMSs into an existing multimodal transportation network, particularly in terms of travel decisions, network congestion, emissions, and the distribution of transportation benefits. The second objective is to explore optimization strategies that improve the overall performance of the multimodal system, particularly when SMSs are operated by private, profit-oriented actors. This involves designing pricing and incentive mechanisms that align commercial interests with public policy goals.

Despite extensive research in transportation modeling, significant limitations persist in representing such complex interactions. Many existing models operate under static assumptions or focus on isolated components of the transportation system, often neglecting the system dynamics, intermodal interactions, and the strategic behavior of service providers. Furthermore, the heterogeneity of users remains underrepresented in many modeling frameworks. Additionally, although dynamic pricing mechanisms have been investigated for SMSs, these studies often abstract away from the broader multimodal system-level context in which SMSs operate \citep{oh_assessing_2020}.  
Similarly, research on incentivization policies, particularly those promoting sustainable travel patterns, remains sparse. In particular, there is a lack of methodological approaches that optimize such policies within multimodal networks. Furthermore, traditional optimization methods often struggle to manage systems characterized by sequential decision-making. In this context, reinforcement learning (RL) has emerged as a powerful approach for developing adaptive policies through interaction with dynamic environments \citep{li_deep_2023}. RL is especially well-suited to transportation systems because it enables the learning of strategies that account for long-term effects, system feedback, and the interplay between multiple stakeholders \citep{qin_reinforcement_2025}.

To address these research gaps, we propose a methodological framework that integrates reinforcement learning with a multimodal macroscopic simulation model. Specifically, we employ an RL-based approach to optimize the decision-making of both service providers and public authorities: the former dynamically adjust pricing strategies to balance supply and demand, while the latter deploy incentive schemes for public transport to promote socially and environmentally desirable travel behaviors. These RL agents interact within an environment that captures multimodal system dynamics, user heterogeneity, and mode competition. 
Through this approach, we aim to investigate how public authorities and private SMSs providers can jointly use dynamic incentives and pricing to improve efficiency, equity, and emissions outcomes, while maintaining the provider’s profitability.

The remainder of the paper is organized as follows. Section \ref{LR} reviews related work from the existing literature. Section \ref{methodology} presents the multimodal modeling framework, with RL-based dynamic pricing and incentivization strategies. Numerical results are discussed in Section \ref{exp_res}. Finally, Section \ref{conclusion} offers concluding remarks and perspectives.

\section{Literature review}
\label{LR}
In what follows, we summarize the most relevant contributions from four key perspectives: (a) modeling approaches for multimodal transportation systems; (b) dynamic pricing approaches designed to maximize profits for SMSs providers; (c) incentivization strategies aimed at optimizing overall system performance; and (d) definitions and measurement approaches of emissions and equity.

\subsection{Multimodal network modeling approaches}
Accurate modeling of multimodal transportation networks is essential for understanding users' choices and estimating system performance. Various works of the literature adopted a time-independent and analytical perspective, primarily focusing on equilibrium conditions. For example, \cite{di_unified_2019}, \cite{li_path-based_2020}, and \cite{wang_convex_2021} proposed carpooling equilibrium frameworks to represent interactions between personal vehicles and SMSs. However, these models exclude public transport (PT) in the modeling. {\cite{zardini_game_2021} proposes a Stackelberg model to represent interactions between public authorities and SMSs providers. However, their approach does not account for user heterogeneity and intermodal behavior.} \cite{ma_economic_2023} explored mode choice and optimal operational decisions for PT and e-hailing operators, while considering the heterogeneity of users' value of time. {In \cite{anzenhofer_potential_2025}, an optimization-based modeling framework is proposed for SMSs to evaluate how regulatory constraints imposed by public authorities influence system-level performance}. A common limitation across these works is their failure to address intermodality, as they do not model trips involving combinations of different modes.

Other works addressed a broader range of travel options, user heterogeneity, and intermodality. For example, \cite{zhu_analysis_2020} examined the impact of ridesharing on PT ridership. \cite{wang_designing_2022} expanded this scope by integrating e-hailing, bikesharing, and PT. \cite{gao_integrated_2025} considered an integrated framework including e-hailing, bikesharing, and PT. They studied the impact of SMSs regulations on different performance metrics. Other studies (e.g. \cite{du_modeling_2022, liu_integrating_2023, kadem_analytical_2024}) have investigated the impacts of integrating PT with SMSs on modal shifts and system performance, though they do not incorporate user heterogeneity.

While these advanced static models significantly improve the representation of user heterogeneity and intermodality, they are inherently limited by their static nature and fail to capture dynamic interactions.

Dynamic models provide a more accurate approach to simulating transportation systems. Some works focused on optimizing SMSs operations, particularly in terms of matching and repositioning, without consideration of their interactions with other modes (e.g., \cite{ma_dynamic_2019}, \cite{alisoltani_can_2021}, \cite{feng_coordinating_2022},  \cite{bujak_ride-pooling_2024}, \cite{homsi_rolling_2024}, \cite{sun_critical_2025}, \cite{de_ruijter_day--day_2025}). A more comprehensive view is adopted by \cite{pi_general_2019}, who formulated a dynamic analytical equilibrium model to analyze interactions within multimodal systems. 

Building on these analytical approaches, simulation-based models offer greater flexibility to represent complex operational dynamics and traveler behavior. For example, \cite{wei_modeling_2020} used Macroscopic Fundamental Diagrams (MFD) to study day-to-day dynamics of traveler choices and traffic conditions, though their model lacked user heterogeneity and intermodality. {\cite{dandl_regulating_2021} proposed a tri-level simulation and optimization-based framework to regulate shared automated mobility on demand services (AMOD) by jointly modeling the interactions between public authorities, service providers, and travelers.} \cite{lorente_using_2025} also proposed a simulation-optimization-based approach to evaluate the environmental impact of ridesharing within intermodal transport systems. \cite{basu_automated_2018} and \cite{oh_assessing_2020} examined the impacts of AMOD using an agent-based simulator (SimMobility). \cite{narayan_does_2019} and \cite{becker_assessing_2020} used MATSim, a multi-agent simulation tool, to investigate the potential of ridesourcing in absorbing private car and PT demand. \cite{lee_assessing_2025} investigated the effects of rebalancing strategies on the resilience of transportation systems, employing a multimodal simulation framework (MnMS) based on MFD.

While these studies have addressed SMSs-PT intermodality and user heterogeneity within dynamic models, their approaches are often intricate and require substantial modeling effort to extend or adapt. In contrast, we propose a macroscopic modeling framework that remains comprehensive yet tractable, offering a flexible foundation that can be more easily adapted to diverse planning and policy objectives.

\subsection{Dynamic pricing for ridesharing}

A significant body of research has addressed pricing strategies for ridesharing services, primarily focusing on maximizing service providers’ profits. Many of these works rely on analytical and optimization-based models. For instance, several studies targeted spatial pricing (e.g., \cite{cachon_role_2017}, \cite{qiu_dynamic_2018}, \cite{zha_geometric_2018}, \cite{bimpikis_spatial_2019}) and temporal pricing (e.g., \cite{qian_time--day_2017}, \cite{zha_surge_2018}, \cite{nourinejad_ride-sourcing_2020}, \cite{yang_integrated_2020}) under varying assumptions of market behavior. These models can examine temporal and spatial pricing separately but not simultaneously. Other works propose spatio-temporal pricing schemes. \cite{asghari_adapt-pricing_2018} developed an optimization-based predictive pricing approach that dynamically adjusts the price based on the current level of supply and demand, as well as the predicted demand. \cite{he_spatio-temporal_2019} also used demand-supply prediction to optimize travel fares to maximize profit. \cite{ma_spatio-temporal_2020} used linear programming to design a spatio-temporal pricing method to maximize social welfare. These approaches usually assume an explicit mathematical relationship between price, demand, and supply.

However, capturing the complex and often stochastic interaction between pricing decisions and the evolving supply-demand balance presents significant challenges for analytical models. Reinforcement learning (RL) has thus emerged as a promising approach to overcome these limitations \citep{qin_reinforcement_2025}. RL frameworks allow the system to learn pricing strategies through interaction with a dynamic environment that encapsulates both endogenous factors (e.g., demand fluctuations, vehicle availability) and exogenous factors (e.g., traffic conditions, policy changes).

In \cite{song_application_2020}, an RL agent is used to represent a taxi provider that applies surge pricing for profit maximization. \cite{huang_deep_2022} applied deep RL to determine dynamic origin-destination fares that maximize profit and the response rate to orders. \cite{chen_spatial-temporal_2021} and \cite{lei_scalable_2023} propose a centralized RL approach to define an origin-based price factor that maximizes system profit, considering the costs of riders and drivers. \cite{turan_dynamic_2020} assume electric autonomous vehicles and incorporate charging decisions into the joint pricing and repositioning problem. Similarly, \cite{li_learning_2024} considered a fleet of autonomous vehicles and proposed a joint deep RL approach for pricing and repositioning. Multi-agent RL has also been investigated to handle coordinated pricing among multiple service providers \citep{sun_dynamic_2024} and zone-based pricing \citep{ge_r2pricing_2025}.

While these studies demonstrate the potential of RL for dynamic pricing, they are often developed in isolation from broader multimodal transportation systems and do not fully account for the interdependencies between pricing, user behavior, and network dynamics. In this study, we aim to bridge this gap by developing a framework that explicitly captures these interactions within a multimodal context.

\subsection{Incentivization strategies}

Beyond optimizing for profitability, diverse incentivization strategies have emerged to guide travelers toward more sustainable options. In \cite{li_incentive-based_2018}, incentives were offered to encourage passengers to shift their departure times to off-peak hours, to minimize the operational cost of the railway service. \cite{zhu_personalized_2020} propose a personalized incentivization framework to promote sustainable travel behaviors. \cite{xiao_optimizing_2021} focus on incentive budget allocation under static equilibrium assumptions through an optimization-based formulation. \cite{zhang_day--day_2021} addressed the car-sharing service by incentivizing drivers to relocate the vehicles to low-supply zones. \cite{wang_integrating_2021} proposed an incentive-based mechanism, using a bi-level optimization model, to encourage demand-responsive transit passengers to modify their pickup and drop-off locations. \cite{sun_evaluating_2022} present a rolling-horizon approach, in which personalized incentives are offered to carpooling users to change their departure time and destination stop, in order to reduce emissions. \cite{ke_leveraging_2023} used bi-level programming to propose a static link-based subsidy strategy to influence the routing of ridesharing vehicles to minimize the total travel time of the system. \cite{bian_public_2024} analyzed the impact of PT discounts on ridership and social welfare. 

However, multimodal interactions significantly impact the overall performance of the system. The literature that addresses this aspect remains sparse. For example, \cite{song_incentives_2021} offers spatial subsidies to carpooling users to improve social welfare and reduce traffic congestion through a multimodal optimization-based framework. \cite{lu_efficient_2024} develops a simulation-based approach that aims to improve accessibility by maximizing the number of ridesharing orders without a suitable public transport alternative, thereby minimizing the total travel time. \cite{zhou_transit_2024} used premium pricing schemes to offer free access to alternative modes when delays for public transit exceed a threshold. \cite{gao_regulating_2024} formulate a profit maximization problem using a market equilibrium model to investigate the impacts of ridesharing on spatial equity when a fixed subsidy is offered to transit users with low income levels. Similarly, \cite{gao_synergizing_2024} explore the use of origin-based incentives for bike-sharing users, while investigating the impact on spatial equity. \cite{guo_equity-aware_2025} develop an analytical framework to optimize the timing for implementing fare-free transit and the length of fare-free transit zone under social equity and social welfare regimes. \cite{huang_unveiling_2025} propose a game-theoretic model in which the public transport operator may subsidize travelers who use a shared mobility service to access first or last-mile trips. 


Studies exploring the use of RL for incentivization are limited, and the majority of these focus primarily on ridesharing services. \cite{wu_spatio-temporal_2022} proposed a combined deep RL and optimization-based framework for spatio-temporal incentives for ridesharing to balance demand and supply levels. In \cite{chen_optimization_2025}, temporal PT incentives are offered to heterogeneous users to promote the use of green modes during the peak hour. \cite{schofield_rebalancing_2024} uses an RL agent to offer temporal, origin-based incentives to bike-sharing users to change their departure zone, in order to maximize the service level. 

These studies represent a shift from purely profit-driven models toward approaches that integrate transportation planning objectives. However, a clear opportunity remains to leverage RL's ability to learn from complex multimodal interactions. This enables the joint optimization of profit alongside environmental and social objectives. To the best of our knowledge, our work is the first to tackle this challenge, considering various system performance metrics, including emissions and equity.

\subsection{Emissions and equity measurement}
A wide array of approaches exists for measuring and estimating transportation emissions. The most fundamental models rely on vehicle-specific fuel consumption to estimate total emissions (e.g., \cite{ sukor_carbon_2017, goodchild_analytical_2018}). Other approaches use microscale models. For example, modal emission models (e.g., IVE, MOVES, CMEM) estimate emissions based on speed, acceleration, and engine characteristics. Alternatively, average speed-based models (e.g., COPERT, EMFAC) estimate emissions using aggregated data such as average speeds, vehicle counts, and fleet composition. These models require careful calibration using real-world measurements from roadside monitoring or remote sensing.

Some approaches take a more aggregated view by relating network-level travel times to emissions. For instance, \cite{ma_emission_2017} and \cite{tan_emission_2021} used link travel times to estimate the vehicular emission rate of the system. Although less precise than microscale models, aggregate travel-time-based methods are scalable, data-efficient, and particularly suitable for evaluating large-scale policy scenarios. Detailed reviews of different types of emission models can be found in \cite{zhong_models_2024}.

In this study, we adopt an aggregate travel-time-based measuring approach, consistent with the nature of our available data and the scale of our analysis. This method enables us to capture relative emissions across the network and evaluate system changes without requiring extensive calibration or vehicle-specific data. 

While emissions offer a critical perspective on the environmental consequences of transportation, they represent only one dimension of system performance. Equally important is considering equity to assess the fairness of policy outcomes and ensure that transportation systems remain socially sustainable.

The literature lacks a unified definition of equity, especially in the transportation field. Nevertheless, it is often defined as the fair distribution of transportation impacts, emphasizing the importance of evaluating who receives benefits, who bears costs, and whether those distributions are considered fair and acceptable \citep{litman_evaluating_2017}. We commonly distinguish between social and spatial equity. Social equity acknowledges that individuals differ in terms of needs and resources. Designing for social equity involves prioritizing access for disadvantaged communities, but presents challenges, including how to define disadvantage, how to quantify needs, and how to balance the redistribution of resources at the individual level. Several modeling studies have incorporated social equity (in terms of income levels, gender, race, etc.) in network design, pricing, or resource allocation frameworks (e.g., \cite{foth_towards_2013, yan_fairst_2019, lee_social_2023, pramanik_equity_2023, gao_regulating_2024}). 

Spatial equity concerns the equal distribution of benefits and burdens among regions and communities. It typically focuses on the geographic distribution and accessibility of transportation services across regions. In this context, the key challenge lies in defining what constitutes equal treatment, particularly where urban infrastructure and travel patterns differ widely. Numerous studies have proposed modeling approaches for spatial equity (e.g., \cite{delbosc_using_2011, najmi_equity_2023, gao_synergizing_2024, gao_regulating_2024, wang_planning_2025}).

{In this study, we define equity as the fair distribution of transportation benefits across both geographic regions and population groups. Specifically, we quantify how benefits are distributed over different zones of the network, addressing concerns related to spatial equity. Moreover, we assess how these benefits are shared among individuals of varying income levels, which relates to social equity. Although our framework does not explicitly prioritize giving more resources to lower-income groups, it does allow for the evaluation of equity impacts across both space and socioeconomic class.}

\subsection{Contribution}

In light of the above review, our work seeks to advance the state of the art through the following key contributions:

\begin{itemize}
    \item We propose a dynamic, multi-class, macroscopic modeling framework that integrates SMSs and PT in a multimodal setting. The model captures user heterogeneity, intermodal travel options, and network-wide traffic dynamics using a trip-based macroscopic fundamental diagram (MFD) traffic simulator.
    \item We introduce a dual-agent reinforcement learning framework. The first agent represents an SMS provider, adjusting prices to maximize profits, while the second represents the public authority offering PT incentives.
    \item We conduct a detailed evaluation of the impacts of SMSs on modal split, travelers' costs, spatial equity, and emissions using the Sioux Falls network as a case study.
    \item We evaluate how coordinated dynamic pricing and incentivization strategies can guide the system toward more equitable and efficient outcomes, despite conflicting objectives among key stakeholders.
\end{itemize}

\section{Methodology}
\label{methodology}
A comprehensive framework for modeling and analyzing multimodal transportation systems is depicted in Figure \ref{fig:stakeholders}. This framework integrates key system components, decision-making agents, and performance evaluation mechanisms to capture the complex, dynamic, and interdependent nature of urban mobility environments.

At the core lies the multimodal transportation system, composed of various transport modes. The system is shaped by the interactions of three principal agent types: travelers, shared mobility service providers, and public authorities. Each agent pursues distinct objectives and operates at a different level within the system.

Travelers act as travel cost-minimizing agents, making daily decisions regarding:
\begin{itemize}
    \item Mode choice, where users select among the available mobility options based on criteria such as price, expected travel and waiting times, and comfort.
    \item Route selection, in which travelers choose the path that minimizes their perceived travel costs, given current network conditions.
\end{itemize}

These decisions jointly determine the travel demand of each mode at a given time and location, which recursively impact the system state and travelers’ costs.

SMSs providers (e.g., carpooling and ridesharing platforms) are profit-maximizing agents that operate directly within the multimodal system. They influence and respond to demand through:
\begin{itemize}
    \item Operational optimization, which involves matching passengers with vehicles to maximize satisfied demand while minimizing operational costs.
    \item Pricing strategies, which consist of dynamically adjusting fares to balance supply with demand, maximize revenue, and regulate service accessibility.
\end{itemize}

In this study, we consider two forms of SMSs, namely carpooling and ridesharing. The key difference between these two lies in the role of the driver. In carpooling, the driver is a commuter with personal travel needs who chooses to share their vehicle with other travelers to reduce costs. In contrast, ridesharing is similar to a shared taxi service, where the driver provides transportation as a job, potentially serving multiple passengers simultaneously. Ridesharing is inherently more complex, as it involves coordinating shared trips, which affects the matching process, costs, waiting times, and vehicle management since it involves handling partially occupied vehicles.

Public authorities operate at a higher strategic level. They influence traveler behavior via targeted PT subsidies to promote system-level optimal options. To assess the effectiveness of different strategies, public authorities refer to a set of performance indicators to quantify societal and environmental objectives.

The interactions among travelers, operators, and infrastructure produce a set of endogenous system states that form feedback loops, directly affecting the subsequent decision-making of all actors. Such interactions are crucial to the system's dynamic adaptation. 

To capture the interactions between all stakeholders, the entire system evolves through a rolling horizon scheme, as illustrated by Figure \ref{fig:RH}. The overall planning horizon is discretized into successive time intervals of a given length $\Delta T$. {Within each time interval $[ \tau , \tau + \Delta T]$, demand levels, system conditions, and control actions are assumed to be stationary, and a sequence of operations is performed as follows}: 

\begin{figure}[ht]
    \centering
    \includegraphics[width=0.9\linewidth]{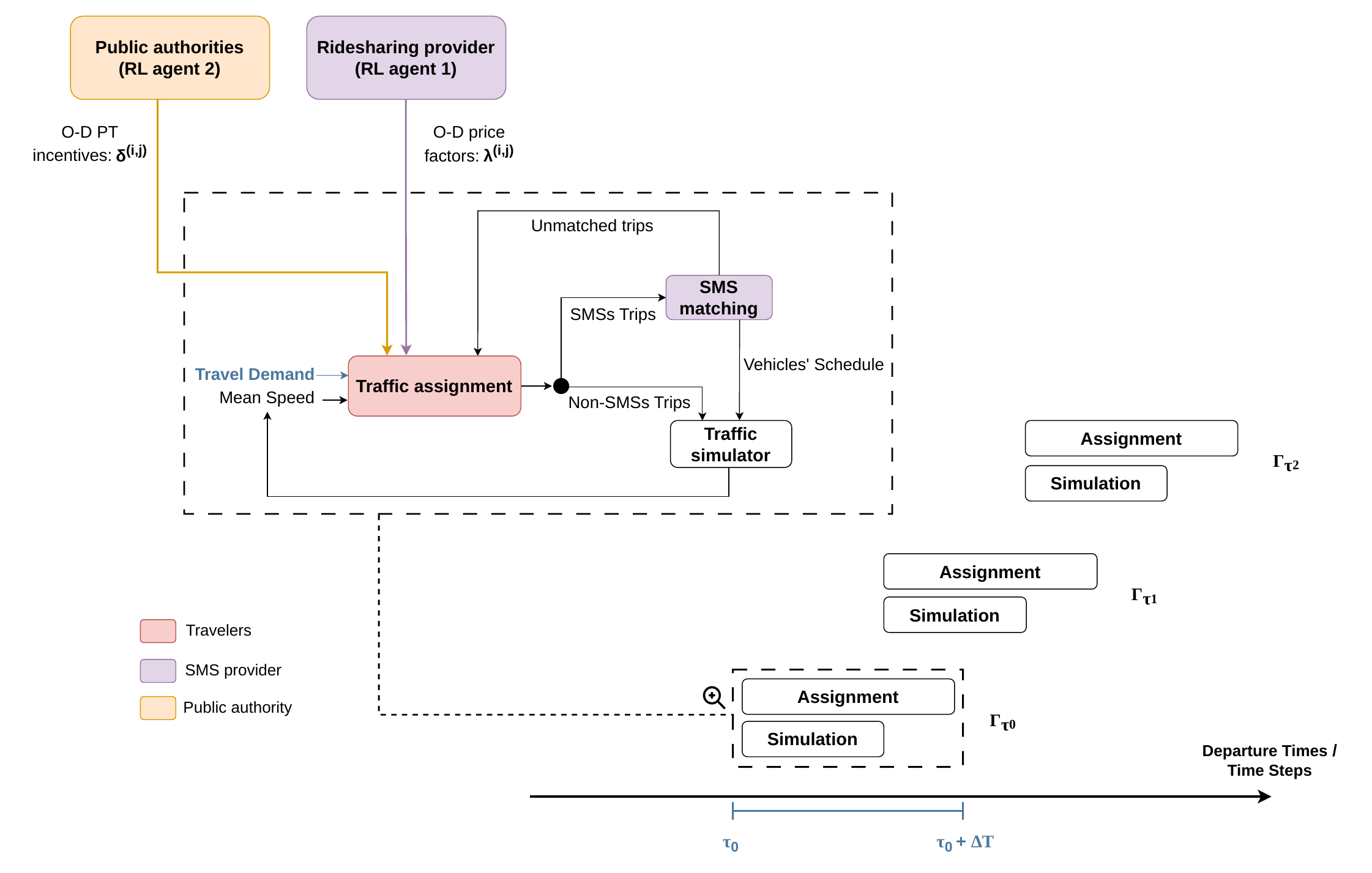}
    \caption{Dynamic Modeling framework for multimodal transportation systems using rolling horizon}
    \label{fig:RH}
\end{figure}
 
\begin{itemize}
    \item Both RL agents observe the current state of the network, and select their respective actions at the beginning of the interval, and those actions remain fixed throughout the interval.
    \item The traffic assignment module allocates travel demand across available modes and paths. To compute travelers’ generalized costs, it uses the fares set by ridesharing platforms and public authorities, and estimates travel times based on average network speeds. 
    \item {The SMSs matching module processes trip requests for SMSs, produced by the UE model, and generates vehicle schedules. Requests that cannot be matched due to vehicle unavailability or time window constraints are reassigned to alternative modes.} {It should be noted that SMSs requests are aggregated over the interval and processed in a batched manner by the matching module. Separate batches are formed for ridesharing and carpooling, and each batch is handled jointly to construct feasible assignments based on the available supply at the beginning of the interval.}
    \item The traffic simulator executes the SMSs schedules, along with other non-shared trips to estimate the real traffic state and outputs average speed and SMSs waiting time values. These outputs feed back into the next iteration, refining travelers’ choices and demand patterns.
\end{itemize}

Iterations continue until convergence, which is achieved when the estimated speeds and waiting times do not change significantly between two consecutive iterations. {Once convergence is reached, the system transitions to the next timestep $\tau + \Delta T$. The traffic state is initialized using the converged values from the previous interval. Ongoing trips that have not yet reached their destination are retained as background traffic and contribute to the congestion in the next interval. Fleet vehicles that are still serving passengers remain unavailable, while idle vehicles become available at their last recorded locations. New demand is then generated for timestep $ \tau + \Delta T$, and both RL agents select new actions based on the updated system state. The assignment-matching-simulation process is subsequently repeated for the new interval.}

{The interaction between the ridesharing provider and the public authority is also modeled within this discrete-time decision process. There is no direct interaction or information exchange between the authority and the SMS provider. Instead, their decisions are coupled indirectly through the transportation system. In particular, at each decision step, the multimodal simulation provides an observation summarizing current system conditions. Each agent then constructs its own state representation from this shared environment. Based on this state, both agents select their actions independently and simultaneously. Their decisions jointly affect travelers’ generalized costs, which in turn influence mode choice, congestion, and service performance. Because both agents act concurrently and influence the same shared environment, each agent’s reward depends not only on its own action but also on the other agent’s policy.}

In this study, we assume that both actors operate on the same time scale. This assumption can be relaxed by allowing public authorities to act on a coarser scale, adjusting incentives less frequently. However, it is worth mentioning that such an approach has a potential drawback: the SMS provider, operating at a higher frequency and with knowledge of the incentive scheme, could anticipate public interventions and adapt its strategy accordingly to maximize its own benefit, thereby strategically exploiting the system.

{It is also worth noting that the pricing and incentivization agents are modeled as independent decision-makers. Although they operate within a shared environment and their actions jointly affect system performance, each agent optimizes its own objective, and learning is conducted independently. Consequently, the resulting policies reflect strategic responses under coupled system dynamics, without relying on an explicit coordination mechanism typically used in cooperative multi-agent RL settings. While this approach enables a tractable and stable representation, extending the framework to fully cooperative multi-agent reinforcement learning constitutes an important direction for future work.}

In the following sections, we describe each component of the framework. A summary of the key notations used throughout the paper is provided in \ref{ap:notation}.

\subsection{Travelers}
\label{mode_choice}

The \textit{Travelers} component represents the decision-making processes of individuals in the transportation system. The core of this module is a multi-class flow-based model that utilizes the user equilibrium (UE) principle. In UE, every traveler tries to minimize their own cost, and equilibrium is reached when no one is willing to change their choices to achieve lower costs \citep{wardrop_theoretical_1952}.

In each time step $\tau$ of the rolling-horizon previously described, a separate assignment problem is formulated. For the sake of simplicity, we present the mathematical formulation of the assignment model within a given time step.

Let us consider an urban transportation network represented as a directed graph $G(E, A)$, wherein links ($A$) represent physical road sections, and nodes ($E$) can represent intersections or zones, depending on the level of network aggregation. A crucial aspect of this model is the consideration of heterogeneous travelers. These travelers are grouped into various user classes, where each class $k$ represents a set of users sharing the same value of time (VOT) $\alpha_k$. We denote $K$ as the set of user classes in the system. Commuters traveling between Origin-Destination (OD) pair $(i,j)$ must simultaneously choose their travel mode and path. The travel options handled by the model are: private vehicles (PV), bus, metro/railway (M), carpooling as a passenger (CP), carpooling as a driver (CD), ridesharing (RS), walking (W) and biking (B); intermodal options between mode $m1$ and $m2$ with a transfer occurring at station (i.e. node) $s$ ($I_s(m1,m2)$) are also possible. Moreover, we take into account the fact that not all travelers have access to a private vehicle. To capture this, we denote by $\zeta_k \in [0,1]$ the proportion of individuals of class $k$ who own a car.

In what follows, we define the generalized costs associated with each travel option. Note that the variables estimated from the simulation outputs are denoted using a bar notation(e.g., $\overline{x}$). 

Let $c_{p,m,k}^{(i,j)}$ denote the generalized cost of path $p$ with travel mode $m$ between OD  pair $(i, j)$ for user class $k$. A path refers to an ordered sequence of network links $a \in A$ that connects a given origin node $i$ to a destination node $j$, representing the route taken by a commuter. 

A commuter who drives alone (PV) is subject to the following cost:

\begin{equation}
    c_{p,PV,k}^{(i,j)} = \alpha_k \cdot \overline{TT_p} + \gamma \cdot L_p + PF_{PV} + \alpha_k \cdot P_{PV}
\end{equation}

\noindent where $\overline{TT_p}$ is the travel time on the path $p$, estimated based on the length of the path and the average speed of the road network. $\gamma$ is the monetary cost per unit of distance, $L_p$ is the length of path $p$. $PF_{PV}$ and $P_{PV}$ denote the parking fare and time, respectively. {Parking time represents the time required to search for and access a parking space at the destination.}

A commuter who participates in carpooling as a driver (CD), and therefore uses their vehicle for the trip, incurs the same cost as a PV, reduced by the fare received from the passenger ($F_{p,m}^{(i,j)}$). In addition, the driver experiences an inconvenience cost ($\beta_m$) which captures the perceived discomfort associated with sharing the vehicle and adjusting routes. The driver also incurs a service time that includes both the matching time and the time associated with passenger pick-up and drop-off ($ST_m$). The generalized cost is given by:

\begin{equation}
    c_{p,CD,k}^{(i,j)} = c_{p,PV,k}^{(i,j)} + \alpha_k \cdot ST_{CD} + \beta_{CD} - F_{p,CD}^{(i,j)}
\end{equation}

The generalized cost for SMSs passengers (CP and RS) is defined as follows, where $\overline{TT_p}$ is the travel time on the path $p$, $\overline{WT_m}$ is the waiting time to be picked up by a vehicle, $ST_m$ is the service time, $F_{p,m}^{(i,j)}$ is the trip fare, and $\beta_m$ is an inconvenience cost reflecting the discomfort of sharing the ride with others. 

\begin{equation}
    c_{p,m,k}^{(i,j)} = \alpha_k \cdot (\overline{TT_p} + \overline{WT_m} + ST_m) + F_{p,m}^{(i,j)} + \beta_m  \qquad m \in [CP, RS]
\end{equation}

Public transport users are subject to an origin-destination fare ($F^{(i,j)}_m$), an inconvenience cost ($\beta_m$), and experience waiting ($WT_m$) and service ($ST_m$) times, where the service time denotes the average duration spent at each stop. The inconvenience cost captures perceived burdens, such as crowding, lack of comfort, or unreliable service. Moreover, we assume a fixed frequency-based setting. Thus, the waiting time for PT passengers depends on their arrival time at the station. For travel time ($TT_m$), we consider a constant speed for the metro, while the speed of buses is estimated from simulation. The generalized cost for both options is as follows:

\begin{align}
    &c_{p,m,k}^{(i,j)} = \alpha_k \cdot (TT_p + \overline{WT_m} + ST_m) + F^{(i,j)}_m + \beta_m  \qquad m=M \\
    &c_{p,m,k}^{(i,j)} = \alpha_k \cdot (\overline{TT_p} + \overline{WT_m} + ST_m) + F^{(i,j)}_m + \beta_m  \qquad m=bus
\end{align}

For soft modes (W, B), the generalized cost only includes a constant travel time ($TT_m$) and inconvenience cost ($\beta_m$), which reflects factors such as physical effort, weather exposure, and perceived safety.

\begin{equation}
    c_{p,m,k}^{(i,j)} = \alpha_k \cdot TT_p + \beta_m  \qquad m \in [W, B]
\end{equation}

With these definitions, the cost of an intermodal option $I_s(m1,m2)$, using modes $m1$ and $m2$ and transferring at station $s$, is the sum of the costs of each mode, given by:

\begin{equation}
    c_{p,m,k}^{(i,j)} = c_{p(m1),m1,k}^{(i,s)} + c_{p(m2),m2,k}^{(s,j)}  \qquad m = I_s(m1,m2) \quad \forall m1, m2 \in \Psi
\end{equation}

Finally, the mathematical formulation of the multi-class UE model is as follows.
\begin{gather}
    \min \; Z = \sum_{k \in K} \sum_{i,j \in E} \sum_{m \in \Psi} \sum_{p \in P_{ij}} a_{p,m,k}^{(i,j)} \cdot \left( c_{p,m,k}^{(i,j)} - c_{k}^{(i,j)*} \right) \label{UE} \\
    \notag
    a_{p,m,k}^{(i,j)} = 
    \begin{cases}
       0  & \text{if }  f_{p,m,k}^{(i,j)} = 0 \\
       1 & \text{if } f_{p,m,k}^{(i,j)} > 0 \\ 
    \end{cases} \\
    \notag
    c_{k}^{(i,j)*} = \min_{p,m} \; c_{p,m,k}^{(i,j)}
\end{gather}

subject to

\begin{align}
    &q_{m,k}^{(i,j)} = \sum_{p \in P_{ij}} f_{p,m,k}^{(i,j)}   \qquad  \qquad \qquad \forall k \in K \; \forall m \in \Psi \; \forall i,j \in E \label{eq3} \\
    &q^{(i,j)} = \sum_{m \in \Psi} \sum_{k \in K}  q_{m,k}^{(i,j)}  \qquad \qquad \ \quad \forall i,j \in E \label{eq4} \\
    &\sum_{i,j \in E} \sum_{k \in K} ( q_{CD,k}^{(i,j)} + q_{I(CD),k}^{(i,j)} ) \leq \sum_{i,j \in E} \sum_{k \in K}  (q_{CP,k}^{(i,j)} + q_{I(CP),k}^{(i,j)}) \label{eq5} \\
    & \sum_{i,j \in E} \sum_{k \in K}  (q_{CP,k}^{(i,j)} + q_{I(CP),k}^{(i,j)}) \leq CAP_{CP} \cdot \sum_{i,j \in E} \sum_{k \in K} ( q_{CD,k}^{(i,j)} + q_{I(CD),k}^{(i,j)} )  \label{eq6} \\
    & y_{RS} \leq \sum_{i,j \in E} \sum_{k \in K} q_{RS,k}^{(i,j)} + q_{I(RS),k}^{(i,j)} \label{eq7}  \\
    & \sum_{i,j \in E} \sum_{k \in K} q_{RS,k}^{(i,j)} + q_{I(RS),k}^{(i,j)} \leq CAP_{RS} \cdot y_{RS} \label{eq8}  \\
    & { \sum_{i,j \in E} q_{PV,k}^{(i,j)} \leq \zeta_k \sum_{i,j \in E} q^{(i,j)} \label{eq9} \qquad  \qquad \qquad  \qquad  \qquad \forall k \in K }\\
    & { x_{a,m} \leq freq_{m,a} \cdot CAP_m \qquad \qquad \; \; \; \forall a \in A, \; \forall m \in \{bus, M\} } \label{eq11} \\
    & { x_{a,m} = \sum_{i,j \in E} \sum_{p \in P_{ij}} f_{p,m,k}^{(i,j)} \cdot \sigma_{a,p,m}^{(i,j)} \quad \quad \; \; \forall a \in A, \; \forall m \in \{bus, M\} } \label{eq12} \\
    &f_{p,m,k}^{(i,j)} \geq 0 \qquad  \qquad \qquad  \qquad  \qquad \forall k \in K \; \forall m \in \Psi \; \forall p \in P_{ij} \; \forall i,j \in E  \label{eq10}  
\end{align}

The objective function minimizes the cost gap between the used options and the one with the minimum cost, thus representing the UE conditions \citep{lin_formulating_2022}. Let $f_{p,m,k}^{(i,j)}$ denote the traffic flow on path $p$ with mode $m$ for users class $k$ between OD pair $(i,j)$. Constraints (\ref{eq3}) and (\ref{eq4}) ensure the flow conservation for each OD pair. 
Capacity constraint (\ref{eq5}) ensures that each carpooling driver has at least one passenger on board. $CAP_m$ represents the maximum passenger capacity for the travel mode $m$. Constraint (\ref{eq6}) ensures that demand can be satisfied and thus, all carpooling passengers can be picked up by a driver. Similarly, let $y_{RS}$ denote the number of available ridesharing service vehicles in the network. Constraints (\ref{eq7}) and (\ref{eq8}) represent capacity constraints for ridesharing, ensuring that supply and demand are well-balanced. {Constraint (\ref{eq9}) ensures that the demand served by private vehicles for each class $k$ satisfies the level of vehicle ownership for class $k$}. {Constraint (\ref{eq11}) bounds the total number of PT passengers on each link by the available supply. Constraint (\ref{eq12}) converts the modal path flows to modal link flows $x_{a,m}$, through the incidence matrix $\sigma$ of the network.} Constraint (\ref{eq10}) is the non-negativity condition on the traffic flow variable. 

{Since the matching process for SMSs is handled externally to the assignment model, capacity constraints (\ref{eq5}-\ref{eq8}) play a preliminary role at the mode choice stage by ensuring a reasonable balance between the potential demand for SMSs and the available vehicle supply. At this level, these constraints prevent the equilibrium formulation from allocating unrealistically large passenger flows to SMSs. However, the framework would remain valid without these capacity constraints in the UE formulation. Including these constraints significantly reduces the number of iterations required between the mode choice and matching stages by limiting excessive demand at the equilibrium level.}

To avoid fractional flows on individual paths, which would create inconsistencies in the subsequent simulation, we model all flow variables as integers. Path-usage indicators $a_{p,m,k}$ are introduced as binary variables and are linked to the corresponding flows through standard big-M constraints, where M is set equal to the total travel demand in this study. We solve this assignment problem using the Gurobi solver \citep{gurobi_optimization_gurobi_2023}.

{It should be noted that this stage is where the effects of pricing and incentivization materialize, as travelers adjust their mode and route choices in response to generalized travel costs. These costs incorporate both monetary components and travel time, thereby linking policy decisions to user behavior. As a result, pricing and incentivization decisions can influence commuter choices both directly, through changes in the monetary cost of specific modes, and indirectly, through its impact on congestion and network conditions, which in turn affect travel times.}

\subsection{Public authorities}
\label{incentivization}

Public authorities constitute a key decision-making entity within the multimodal transportation system. In the context of this study, their role focuses primarily on managing public transport services and deploying dynamic incentives to guide commuter choices toward socially and environmentally beneficial outcomes.

The decision variable of the public authorities is the spatio-temporal dynamic allocation of monetary incentives to PT users, including those using intermodality with SMSs. These incentives take the form of fare reductions and are defined on an origin-destination basis. Consequently, the actual fare paid by a PT passenger traveling from node $i$ to node $j$ at time-step $\tau$ is given by: 
\begin{equation}
    F^{(i,j)}_m (\tau) = b_m - \delta^{(i,j)} (\tau) \qquad m \in [bus, M]
\end{equation}
Where $b_m$ denotes the fixed fare of public transport and $\delta^{(i,j)} (\tau)$ is the O-D incentive given by authorities at time step $\tau$. 

These incentives are determined by a deep reinforcement learning agent representing the public authority defined by its state, action, and reward functions:

\begin{itemize}
    \item \textbf{State}: Consider $p_{ij}^\tau$ as the PT demand between the OD pair $(i,j)$ at time $\tau$. The state for the authorities agent includes the level of PT demand departing from a node $n$ ($\sum_{j \in E} p_{nj}^\tau$), the level of PT demand arriving at node $n$ ($\sum_{i \in E} p_{in}^\tau$), time of the day $\tau$, and available budget $B_\tau $.
    
     \item \textbf{Action}: The action space consists of providing to PT users fare reductions for each OD pair $(i,j)$. However, using a different incentive for every OD pair leads to an action space of size $N^2$, where $N$ is the number of nodes in the network, making the problem computationally intractable for large $N$. 
     
     To address this, we reduce the dimensionality of the action space by defining each OD-based incentive $\delta^{(i,j)} (\tau)$ as the average of two node-specific values: an origin-based incentive $\delta^i_o (\tau)$ and a destination-based incentive $\delta^j_d (\tau)$. That is, 
     \begin{equation}
         \delta^{(i,j)} (\tau) = \frac{1}{2}(\delta^i_o (\tau) + \delta^j_d (\tau)) \qquad \delta^i_o (\tau), \delta^j_d (\tau) \in I
     \end{equation}
     where $I$ is the set of predefined incentives.
     
    Under this formulation, the RL agent only needs to determine two incentives per node, resulting in a significantly smaller action space of size $2N$. Figure \ref{fig:RL_incen} depicts this space reduction strategy. {Potential boundary issues induced by this decomposition strategy are discussed in \ref{ap:BI}}.

    \begin{figure}[ht]
        \centering
        \includegraphics[width=.8\linewidth]{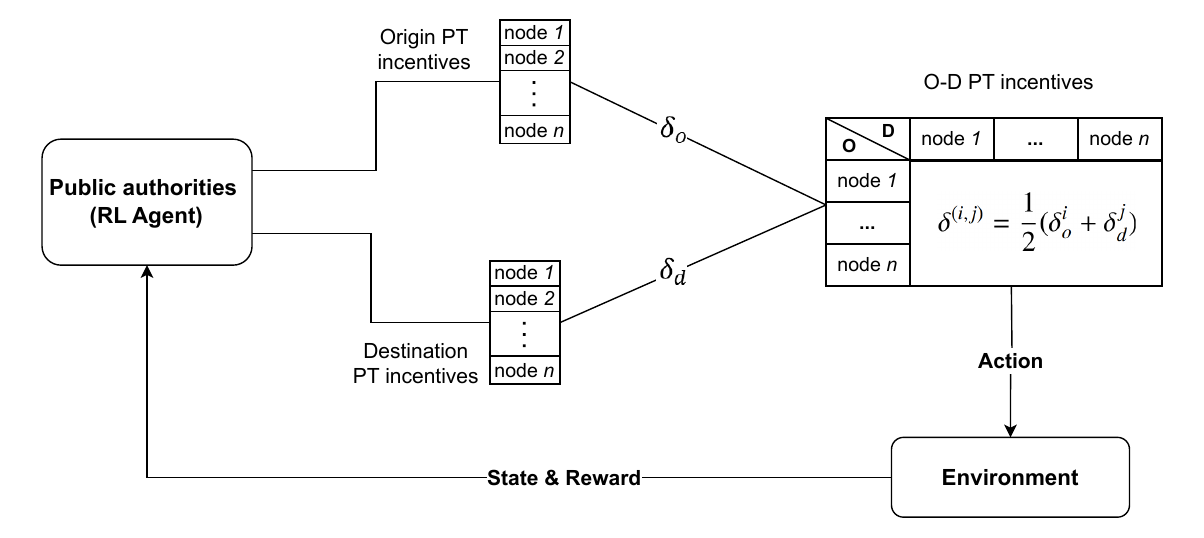}
        \caption{RL-based dynamic incentivization strategy by public authorities}
        \label{fig:RL_incen}
    \end{figure}

    \item \textbf{Reward}: The objective of authorities is to optimize overall system performance. In this study, we define the system performance with three indicators: 
    \begin{itemize}
        \item \textbf{Total Generalized Cost (TGC):} We aggregate the individual generalized travel costs ($c_{p,m,k}^{(i,j)}$) for all commuters across the whole transportation network. The mathematical formulation is as follows: 
        \begin{equation}
            TGC = \sum_{k \in K} \sum_{i,j \in E} \sum_{p \in P_{ij}} \sum_{m \in \Psi} c_{p,m,k}^{(i,j)} \times f_{p,m,k}^{(i,j)}
        \end{equation}
    
        \item \textbf{Equity:} Equity is evaluated by analyzing how equally transportation benefits are distributed among different commuter classes and geographical regions. {In this study, the notion of benefit refers to the relative improvement in generalized travel cost compared to a baseline scenario. Since generalized cost includes both time and monetary components, realized benefits are endogenously determined by the system dynamics and user responses, and are not necessarily proportional to the level of incentives received.} To quantify this, we consider the Gini index, where a low value indicates good spatial equity and a similar level of benefits, while a high Gini index suggests uneven benefits and poor spatial equity \citep{gao_synergizing_2024}. Consider the proportion of users of class $k$ departing from node $i$, denoted as $Q_k^i$, and the average benefits from a policy or scenario $S_1$ relative to a baseline scenario $S_2$ for class $k$ users departing from node $i$, denoted as $U(S_1, S_2)_k^i$. The formula for the Gini index is:
        \begin{equation}
            G = \frac{\sum_{i \in E} \sum_{j \in E} \sum_{k \in K} \sum_{l \in K}  Q_{k}^i Q_{l}^j | U(S_1, S_2)_{k}^i - U(S_1, S_2)_{l}^j |}{2 \cdot (\sum_{i \in E} \sum_{k \in K} Q_{k}^i)^2 \cdot U(S_1, S_2) }
        \end{equation}
        where 
        \begin{equation}
            U(S_1, S_2) = \frac{\sum_{i \in E} \sum_{k \in K} Q_{k}^i \cdot U(S_1, S_2)_{k}^i }{\sum_{i \in E} \sum_{k \in K} Q_{k}^i}
        \end{equation}
        Other metrics can be used to measure equity in a transportation network, each providing a particular insight. In this work, we use the Gini index as a reward function during training, while evaluating the model across various metrics defined in \ref{ap:metrics}. 

        {In the present framework, we primarily focus on spatial equity, as incentivization decisions are defined over space and time. In addition, we analyze the distribution of benefits across user groups to provide complementary insights into social disparities. While alternative formulations could explicitly target specific groups, the adopted approach allows us to evaluate equity outcomes in a consistent and system-driven manner. Extending the framework to include explicit social equity objectives constitutes an important direction for future work.}
        
        \item \textbf{Emissions:} This metric quantifies the environmental impact of the transportation system. In this work, we adopt a well-established vehicular emission model that describes a link-based emission rate as a nonlinear function of link travel time and length \citep{tan_emission_2021, ma_emission_2017}. We denote $e_{a}$ as the emission rate of link $a$, expressed in grams per vehicle per hour (g.v.h). To obtain a network-wide indicator, we aggregate these individual link emission rates into a System Emissions Rate (SER), calculated as follows: 
        \begin{align}
            SER & = \sum_{a \in A} e_{a} \nonumber \\
            & = \sum_{a \in A} 0.2038 \cdot tt_{a} \exp({\frac{0.7962 \cdot l_{a}}{tt_{a}}}) \cdot x_a
        \end{align}
        Here, SER is expressed in grams per hour (g.h), $tt_{a}$ represents the travel time on link $a$ in minutes, $l_a$ is the length of link $a$ in kilometers, and $x_a$ is the vehicular flow of link $a$.

        {It should be noted that emissions from bus operations are included in the SER calculation, as buses are considered in the link flow variable $x_a$. In contrast, metro emissions are not explicitly considered, as metro systems operate on dedicated infrastructure and their emissions are less sensitive to short-term demand variations and pricing decisions.}
    \end{itemize}

    {In this study, the public authority agent is trained using a single-objective formulation, resulting in three separate RL policies that respectively optimize TGC, SER, and G. This design choice allows the analysis to isolate the effects of different policy priorities on system outcomes without imposing an arbitrary weighting scheme across potentially conflicting objectives. From a policy perspective, the three learned policies should be interpreted as decision-support benchmarks, enabling policymakers to evaluate system-wide consequences associated with prioritizing different regulatory goals.} 

\end{itemize}

\subsection{Shared mobility service providers}
\label{SMSprovider}
Within our modeling framework, SMSs are represented by two distinct providers: one operating a carpooling service and the other a ridesharing service. This separation reflects the fundamental operational and managerial differences between the two services. Treating them as independent entities allows us to capture their unique decision-making processes and their competitive interaction in terms of modal split. Each provider faces a specific optimization problem shaped by its service structure and operational constraints.

Both providers engage in a matching process that pairs passenger requests with available vehicles. {The same matching algorithm, presented in \ref{ap:matching}, is used for both ridesharing and carpooling: it is applied to distinct pools and executed separately and in parallel.}

\subsubsection{Capooling provider}

For the carpooling provider, the central challenge lies in driver availability, and the influence of carpooling on system dynamics is primarily determined by the number of such drivers and their geographical location. 

Fares are distance-based and static. The trip fare paid by a carpooling passenger, using path $p$ between an OD pair $(i,j)$ is formulated as follows: 
\begin{equation}
    F_{p,CP}^{(i,j)} =  b_{CP} \times L_p  
\end{equation}

where $b_{CP}$ is the base fare per unit of distance and $L_p$ is the length of path $p$. 

It is important to note that several dimensions of carpooling operations are intentionally simplified in this study. First, dynamic pricing is not modeled. Unlike the ridesharing provider, which employs an RL-based pricing strategy, the carpooling fare structure is kept fixed. This reflects the practical reality of most carpooling platforms, where the voluntary, peer-to-peer nature of the service makes dynamic price signals less actionable.

Second, dynamic compensation schemes for drivers are not considered. Instead, the fare paid by each passenger goes directly and entirely to the driver as revenue. Under this model, the driver’s financial gain scales naturally with the number of passengers sharing the vehicle. This provides an endogenous incentive for drivers to accept shared rides, without requiring a centralized compensation mechanism. More sophisticated driver reward structures could be incorporated in future extensions.

\subsubsection{Ridesharing provider}
\label{pricing}

{The ridesharing provider operates with a fleet of professional drivers under a centralized management strategy, where drivers comply with the platform’s dispatching and scheduling decisions. Strategic participation behavior and driver-initiated repositioning decisions are therefore not modeled in this study. While this assumption enables a clearer analysis of system-level interactions, extending the framework to incorporate such behavioral dynamics constitutes an important direction for future work.}

The ridesharing provider employs dynamic spatio-temporal pricing to adjust price factors in real-time. Specifically, the trip fare paid by an RS passenger using path $p$ between an OD pair $(i, j)$ and departing at time $\tau$ is:
\begin{equation}
    F_{p,RS}^{(i,j)} (\tau) = \lambda^{(i,j)} (\tau) \times b_{RS} \times L_p  
\end{equation}
where $b_{RS}$ is the base fare per unit of distance and $\lambda^{(i,j)} (\tau)$ is an origin-destination price factor defined by the ridesharing provider at time-step $\tau$. 

The dynamic pricing strategy relies on a deep reinforcement learning agent that dynamically adjusts trip fares to maximize profits in response to fluctuating demand and supply conditions across the network. The RL agent is defined as follows:   

\begin{itemize}
    \item \textbf{State}: The state observed by the service provider agent comprises real-time information related to ridesharing demand and supply. Consider $d_{ij}^\tau$ as the level of ridesharing demand between $i$ and $j$ at time-step $\tau$. The state representation of this RL agent, for each time step $\tau$, includes the level of ridesharing demand departing from a node $n$ ($\sum_{j \in E} d_{nj}^\tau$), the level of ridesharing demand arriving at node $n$ ($\sum_{i \in E} d_{in}^\tau$), the number of vehicles available at node $n$ ($v_n^\tau$), and the time of the day $\tau$. 

    {The vehicle availability array in the state representation provides an entry point for incorporating strategic driver behavior and imperfect compliance, such as selective acceptance of requests, which can be modeled through stochastic supply responses.}
    
    \item \textbf{Action}: The action space consists of adjusting ridesharing fare levels by applying multipliers $\lambda^{(i,j)} (\tau)$ to the base fares for each OD pair $(i,j)$, based on demand-supply imbalances across nodes and time periods. Similar to the incentivization strategy (see Section \ref{incentivization}), the agent defines two price factors per node: one for the origin node $\lambda^i_o (\tau) \in ]0,2]$ and the other for the destination node $\lambda^j_d (\tau) \in ]0,2]$. The resulting O-D price factor $\lambda^{(i,j)} (\tau)$ is computed as the average of these two values:
    \begin{equation}
        \lambda^{(i,j)} (\tau) = \frac{1}{2}(\lambda^i_o (\tau) + \lambda^j_d (\tau))
    \end{equation}
    
    \item \textbf{Reward}: The reward for the ridesharing provider is its profit, which is derived from the revenue generated by ridesharing trips minus operational costs, defined as:
    \begin{equation}
    \label{eq:RSreward}
        R_t = \sum_{i,j \in E} d_{ij}^\tau \sum_{p \in P_{ij}} F_{p,RS}^{(i,j)} (\tau) - o_{ij}
    \end{equation}
     where $d_{ij}^\tau$ as the level of ridesharing served demand between $i$ and $j$ at time-step $\tau$, $o_{ij}$ is the cost of operating ridesharing vehicles between an OD pair $(i,j)$. Recall that $F_{p,RS}^{(i,j)} (\tau)$ is the trip fare for ridesharing passengers using path $p$ between an OD pair $(i,j)$ and departing at time $\tau$. 
\end{itemize}

\subsection{Traffic simulator}
\label{simu}
\begin{figure}[ht]
    \centering
    \includegraphics[width=0.9\linewidth]{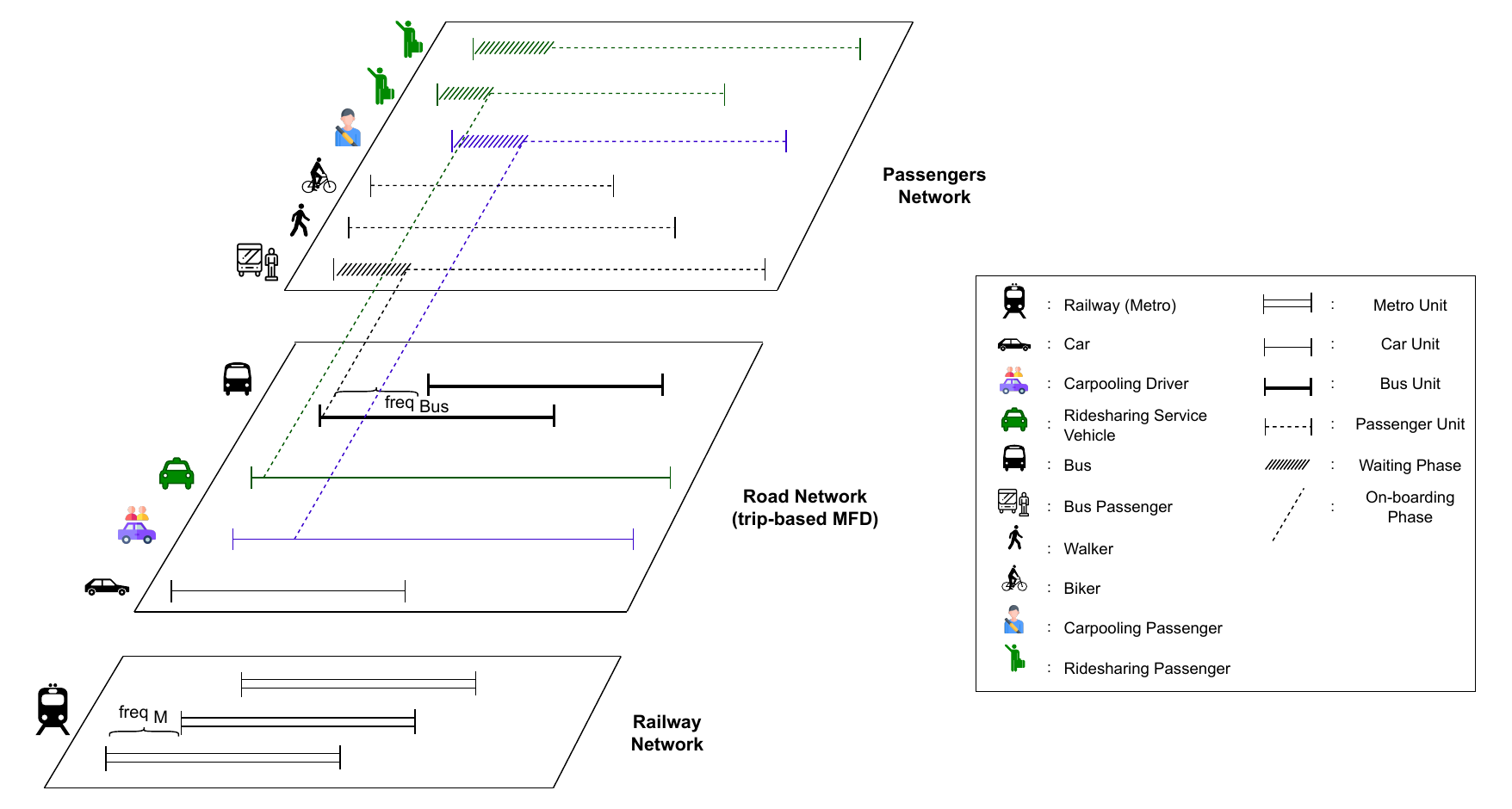}
    \caption{Multimodal trip-based macroscopic traffic simulator}
    \label{fig:MFD}
\end{figure}

To simulate the transportation network, we adopted a trip-based multimodal macroscopic simulation approach, as illustrated in Figure \ref{fig:MFD}. Since commuters have different OD pairs and follow different routes, their trip lengths vary depending on the selected paths. Therefore, mode and path choices are mapped to corresponding trip lengths using the transportation network graph.

The road network is modeled with a trip-based MFD that represents individual trajectories while maintaining a simplified description of network dynamics \citep{lamotte_morning_2018, mariotte_macroscopic_2017}. It captures the dynamics of private vehicles, carpooling drivers, ridesharing service vehicles, and buses. Trips with a passenger role (RS, CP, bus, M) are not considered in the accumulation of the road network and are modeled in a separate layer. The waiting time for SMSs passengers is the time between their departure time and the actual pickup, and their travel time is linked with the assigned vehicle and determined by the MFD.

Because intermodal trips are represented as a sequence of distinct modal trips, the commuter's status may evolve during the simulation. For example, a carpooling driver, initially traveling within the road network under MFD-based speed dynamics, may park at a station and continue the journey as a metro passenger, now simulated within the passenger network using a fixed metro speed.

Public transport is modeled at an aggregate level using a frequency-based representation. Rather than modeling explicit PT schedules with individual lines, stops, and departure times, we assume that PT vehicles enter the network at fixed intervals. Specifically, every $freq_m$ time units, a public transport vehicle of mode $m$ (e.g., bus or metro) is introduced into the system. This abstraction enables a tractable and scalable simulation of PT dynamics, eliminating the complexity of route-level timetables.

A distinction is made between metro/railway and bus services. The former operates on dedicated infrastructure with a constant, exogenously defined speed, and is thus unaffected by road traffic conditions. In contrast, buses use the road infrastructure and are consequently impacted by congestion. We acknowledge that this assumption may not hold in all urban contexts, particularly in cities with dedicated bus lanes. In such cases, the bus system can be represented similarly to the metro mode with a constant speed.

The waiting time for PT passengers is the time between their departure time and the departure time of the next PT unit. {Note that at the simulation level, PT vehicles are not capacity-constrained, meaning that crowding effects and bus-full scenarios are not dynamically propagated. Addressing this simplifying assumption would require extending the simulation layer with a crowding-based disutility function, which we leave for future work.}


\section{Numerical experiments}
\label{exp_res}

This section presents the numerical experiments conducted to assess the applicability of the proposed framework in a multimodal urban setting. We first introduce the test case that provides the experimental environment. We then analyze the effects of various incentivization strategies to demonstrate their potential in enhancing the overall performance of the system. In particular, we compare the outcomes of dynamic incentivization, where subsidies are adaptively adjusted through reinforcement learning, with those of static subsidy schemes. Finally, we extend the analysis to  include a profit-maximizing ridesharing provider. In this joint configuration, dynamic pricing and dynamic incentivization are simultaneously deployed, allowing the agents to learn how to balance their conflicting objectives. The analysis evaluates the resulting trade-offs across efficiency, equity, environmental, and financial dimensions.

\subsection{Test case}
\begin{figure}[h]
    \centering
    \includegraphics[width=0.7\linewidth]{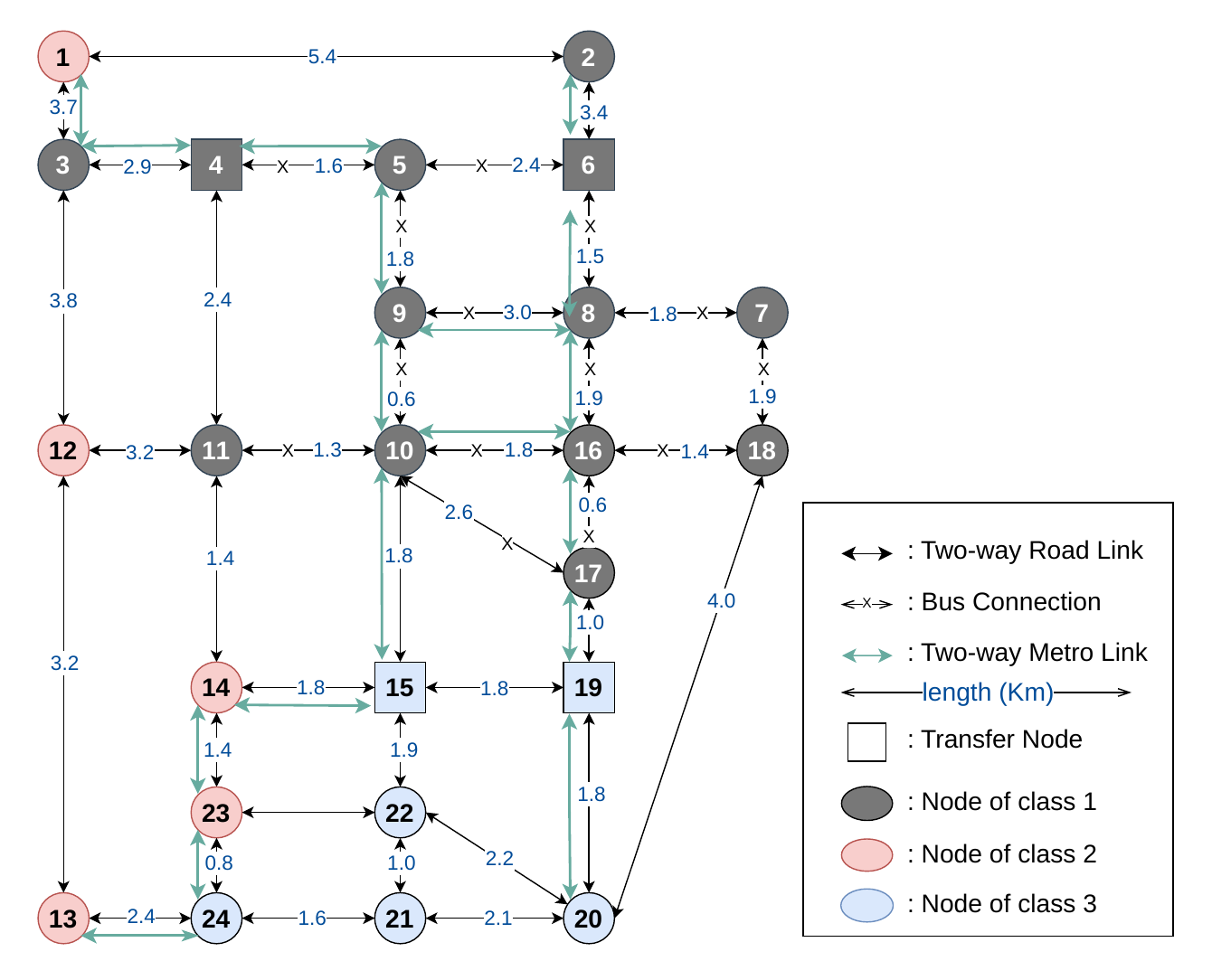}
    \caption{Sioux Falls network with PT infrastructure and heterogeneous user classes}
    \label{fig:SF}
\end{figure}

To evaluate the performance and applicability of the proposed framework, we conduct numerical experiments using the Sioux Falls network, a well-established benchmark in transportation modeling, as shown in Figure \ref{fig:SF}.

The road network topology and OD travel demand matrix are derived from the dataset presented in \cite{stabler_transportation_2016}.
Furthermore, as the original OD data do not specify temporal departure patterns, we generate the departure times of travelers stochastically for the morning peak period (i.e., from 7:00 to 10:00 am). We use a similar distribution as the one proposed by \cite{chakirov_enriched_2014} for this time period. The public transport infrastructure is adapted from \cite{yin_simulation-based_2022}. It includes a metro configuration, modeled as separate links, and a set of bus stops operating on the shared road network. 

{We consider three user classes, each characterized by a different VOT, to reflect income-based heterogeneity across the population. In other words, the whole travel demand is divided into three classes: class 1 represents low VOT (i.e., low-income) users, while class 3 represents high VOT users.} The class assignment is determined based on the average income level associated with each origin node, calibrated from open-source socioeconomic data\footnote{https://zipatlas.com/us/sd/sioux-falls/zip-code-comparison/highest-median-household-income.htm}. Figure \ref{fig:SF} also illustrates the spatial distribution of the user classes, with colors indicating the dominant class within each node. {Moreover, private vehicle ownership is assumed to be homogeneous across user classes in our experiments. In other words, we simplify constraint (\ref{eq5}) of the UE model and assume a single $\zeta = 0.7$ for the entire population.} Other input parameters are introduced in \ref{ap:training}.

\subsection{Impacts of incentivization strategies on system performance}
In this section, we evaluate the system-wide impacts of different incentivization strategies, focusing on the effect on SMSs adoption, system efficiency, and commuters' choices. {The objective is to understand how different policies shape the multimodal system,  illustrate the trade-offs associated with each strategy, and evaluate the potential of dynamic, learning-based approaches in enhancing the overall system performance.} We consider various scenarios:

\begin{itemize}
    \item \textbf{Scenario A - baseline}: standard scenario where SMSs are available, alongside PV and PT, either as door-to-door services or for first/last mile mobility. For this scenario, we don’t consider any specific incentives. This scenario serves as the reference point for evaluating the performance of all policies in the following experiments, unless specified otherwise.
    \item \textbf{Scenario B - SMSs with class-based incentive (Incen. low VOT)}: a static policy where low VOT users (class 1) are granted a pass that gives them a 50\% discount for sustainable travel options (PT and SMSs). {This scenario is presented as an affordability-oriented policy, where incentives are targeted toward cost-sensitive travelers.}
    
    \item \textbf{Scenario C - SMSs with restricted intermodality incentive (Incen. Rest. SMSs+PT)}: an alternative static policy in which a 50\% discount is only granted to intermodal SMSs users to whom PT does not provide an accessible connection between the origin and destination. In other words, commuters pay the full fare for the SMSs trip in cases where a feasible PT path exists between the OD pair. Conversely, when no direct PT alternative is available, a 50\% discount is applied to use SMSs for the first or last mile. {This scenario represents an accessibility-oriented policy, aiming to improve spatial coverage across the network and to encourage integration between SMSs and PT, thereby supporting multimodal travel patterns.}

    \item \textbf{Scenario D - O-D Dynamic PT incentivization (O-D Dyn. Incen.)}: origin-destination based PT incentives are determined dynamically by an RL agent trained to minimize the total generalized cost of the system. {This scenario explores the potential of dynamic, learning-based policy design to balance multiple objectives.}
\end{itemize}

It is important to note that for the static policies, the choice of a 50\% discount is motivated by the fact that such a reduction is where notable shifts in travel behavior start to emerge, especially for SMSs. This can be explained by the relative cost structure of the modes: in the test case, the average PT fare is about 3 USD, while the average SMSs fare is around 10 USD. However, the discount rate remains a critical parameter that requires careful calibration. It is precisely through such calibration that the dynamic incentivization strategy offers a promising approach.

The following subsections present the results for the simulation scenarios across various evaluation dimensions, namely modal split, traffic congestion, vehicular emissions, and equity.

\subsubsection{Impacts on modal split}
Table \ref{tab:mod_share} presents the modal split (i.e., the proportion of trips assigned to each transportation mode) under the different scenarios described previously. For clarity, modes are aggregated into three categories: private vehicles, public transport, and SMSs. The SMSs category includes door-to-door carpooling, and ridesharing, as well as instances where these modes are used for first and last-mile connections. Similarly, the PT category includes bus and metro modes, whether used directly or through intermodality. PV category also accounts for park-and-ride users.

\begin{table}[h!]
\centering
{\small
\begin{tabular}{lcccc}
\hline
Modal share (\%) & \textbf{(A) Baseline} & \textbf{(B) Incen. low VOT} & \textbf{(C) Incen. Rest. SMSs+PT} & \textbf{(D) O-D Dyn. Incen.} \\
\hline
PV  & 39 & 21 & 23 & 32 \\
SMSs & 28 & 57 & 35 & 25 \\
PT  & 49 & 38 & 56  & 58 \\
\hline
\end{tabular}
}
\caption{Evaluation of the modal split under different scenarios. It should be noted that the modal split values reported do not sum to 100\%, as the three categories are not mutually exclusive}
\label{tab:mod_share}
\end{table}

The baseline scenario A establishes the starting point, with public transport representing 49\%, followed by PV at 39\% and SMSs at 28\%. In scenario B, where a subsidy is offered to low VOT users, SMSs adoption increases dramatically to 57\%. This expansion occurs mainly at the expense of PV, which falls to 21\%, suggesting that the policy is effective in reducing private car use. However, PT ridership also declines substantially to 38\%. This outcome reveals a substitution effect: the policy targets a specific user class rather than spatial contexts or modal cooperation, resulting in a decline in PT ridership while shifting demand away from private vehicles toward door-to-door SMSs trips. The scenario demonstrates the strong responsiveness of low VOT travelers to financial incentives. 

Scenario C produces a much more balanced outcome. In this case, PT ridership rises to 56\%, surpassing the baseline, while SMSs maintain a significant share at 35\%. 
Such a policy demonstrates strong potential for enhancing accessibility in urban areas, while avoiding issues related to the competitive side of SMSs. These results also suggest that SMSs can serve as an effective lever to reduce car dependence. However, careful policy design is essential to avoid counterproductive effects, particularly decreases in PT ridership.

Finally, in scenario D, PT achieves its highest share at 58\%, confirming the agent’s ability to favor cost-efficient public transport usage. At the same time, SMSs' share drops to its lowest level at 25\%, while PV reaches 32\%. This increase in PV share, compared to the static incentivization scenarios, can be attributed to the agent’s behavior, as it minimizes the system cost and does not necessarily encourage SMSs use, especially for intermodal connections. Instead, the agent directs travelers toward PT by adjusting the incentives according to their origin and destination. However, the PV share remains less than the baseline scenario, which suggests that some commuters rely on the park-and-ride option more often.

\subsubsection{Impacts on traffic congestion and vehicular emissions}
\begin{figure} [h]
    \centering
    \includegraphics[width=0.75\linewidth]{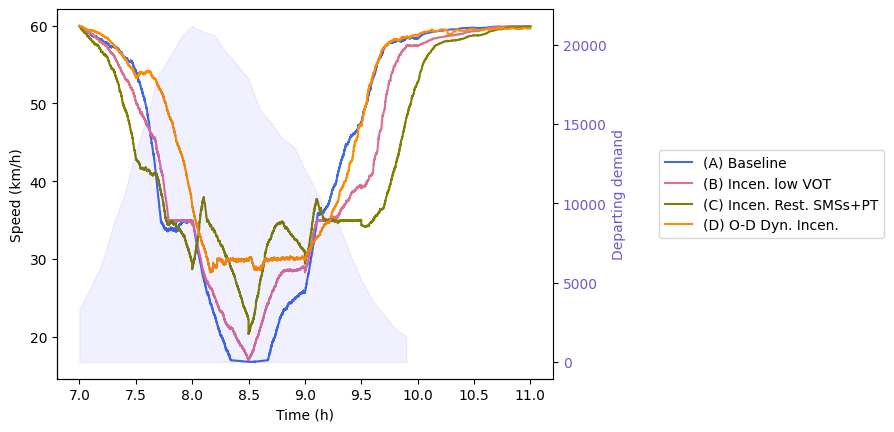}
    \caption{Temporal evolution of average network speed under different scenarios. The shaded area represents the distribution of departure times}
    \label{fig:speed}
\end{figure}

Traffic congestion is assessed using the average network speed observed at each timestep. Figure \ref{fig:speed} illustrates the temporal evolution of average speed across the four scenarios. Additionally, the figure illustrates the distribution of departure times within the population, offering further insight into the temporal patterns of travel demand.

The introduction of subsidies for low-VOT users (scenario B) reduced the number of vehicles on the road, which in turn alleviated congestion during the morning peak. Compared to the baseline, the average network speed was higher at the most congested times, reflecting efficiency gains from fewer cars and more shared trips. However, the improvements were mainly confined to the peak window. Restricting incentives to intermodal SMSs-PT users (scenario C) resulted in a more pronounced impact on congestion during the morning peak, with higher average network speeds between 8:00 and 9:00 compared to both the baseline and scenario B. However, the results also show that the recovery from congestion was slower after, and new congestion points emerged outside the traditional peak window. {This effect is primarily driven by the longer duration of intermodal trips, which keeps vehicles active on the network beyond the peak period and delays congestion recovery. In other words, intermodality incentives helped to relieve pressure during the peak, but mainly by shifting congestion to other times and locations, rather than eliminating it.}

The dynamic incentivization strategy (scenario D) achieved the strongest congestion performance. Overall, it maintained higher network speeds over the time horizon, with particularly good performance at the peak of congestion around 8:30, where it outperformed both static subsidy policies. Although speeds were slightly lower than scenario C at certain moments (around 8:15 and 8:45), scenario D offered a smoother evolution of network conditions, avoiding the sharp fluctuations observed under intermodality incentives. Most importantly, recovery after the peak was faster and more stable, preventing the secondary congestion wave that appeared in scenario C. These results indicate that adaptive incentives not only alleviate peak-hour congestion but also ensure a more consistent network performance throughout the morning period. {To complement this analysis, we provide in \ref{ap:inc_act} a brief analysis of the spatio-temporal distribution of the learned incentive actions in relation to the observed PT demand.}

The environmental implications align with these congestion dynamics. Table \ref{tab:emiss} reports the system emission rate under each scenario. In the baseline, emissions are relatively high (1198Kg.h), consistent with the high reliance on private vehicles. 

{A particularly noteworthy result emerges in scenario B. Although this configuration reduces private vehicle usage, it leads to a marginal increase in system emissions (1201kg.h). This counterintuitive outcome arises from modal reallocation effects: while some private vehicle trips shift toward public transport, a non-negligible share of public transport users is simultaneously attracted toward SMSs. Since shared mobility operates on the road network and contributes to traffic congestion, the average speed profile does not improve sufficiently to avoid this effect, and this rebalancing partially offsets the environmental gains from reduced private car use. This result highlights the risk of shared mobility services cannibalizing public transport ridership when regulatory instruments are not coordinated.}


In contrast, scenario C lowers emissions to 1157Kg.h, mainly by reinforcing PT’s role and alleviating peak congestion. However, the slower recovery observed in the congestion analysis means that parts of the network remain at suboptimal speeds for a longer period, thereby constraining overall environmental gains.

Scenario D achieves the lowest SER (1082Kg.h). By smoothing traffic conditions, preventing secondary peaks, and accelerating the return to free-flow speeds, this policy reduces the system emission rate. Although PV share is higher, the efficiency gains from improved speed conditions dominate, leading to the best environmental performance. This highlights the importance of adaptive strategies that directly manage network speeds, rather than policies that only shift demand between modes.

It is important to note that these results are shaped by the SER formulation used in this study, which relies primarily on average network speeds (and indirectly, trip lengths and the number of cars on the road). Using a more complex metric that also incorporates vehicle occupancy or the degree of sharing could lead to different outcomes. The design of such integrated emission metrics remains an open question in the literature, and future research should explore how best to capture both traffic dynamics and modal efficiency in environmental evaluations.

\begin{table}[h!]
\centering
\begin{tabular}{lcccc}
\hline
 & \textbf{(A) Baseline} & \textbf{(B) Incen. low VOT} & \textbf{(C) Incen. Rest. SMSs+PT} & \textbf{(D) O-D Dyn. Incen.}\\
\hline
SER (Kg.h) & 1198 & 1201 & 1157 & 1082 \\
\hline
\end{tabular}
\caption{Evaluation of the system emission rate under different scenarios}
\label{tab:emiss}
\end{table}

\subsubsection{Impacts on equity}

\begin{figure}[h!]
  \centering
  \begin{subfigure}{0.45\textwidth}
    \includegraphics[width=\linewidth]{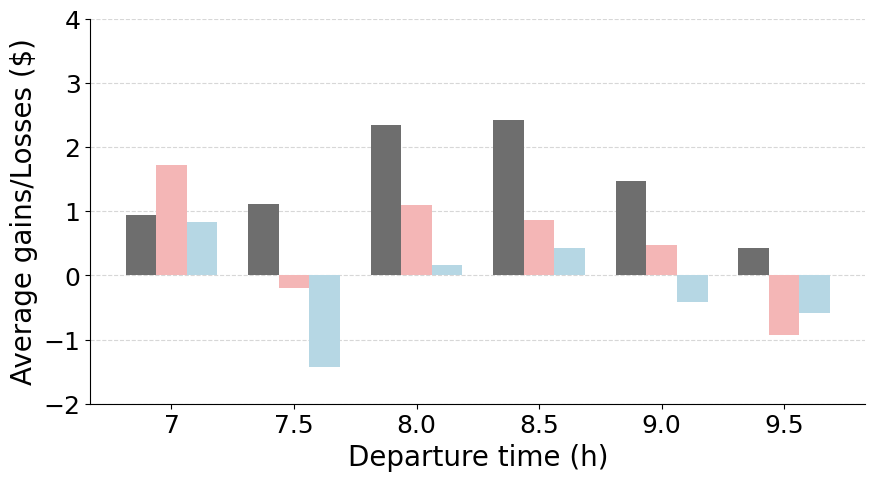}
    \caption{Gains and losses of (B) || (A)}
    \label{fig:gains1}
  \end{subfigure}
  \begin{subfigure}{0.45\textwidth}
    \includegraphics[width=\linewidth]{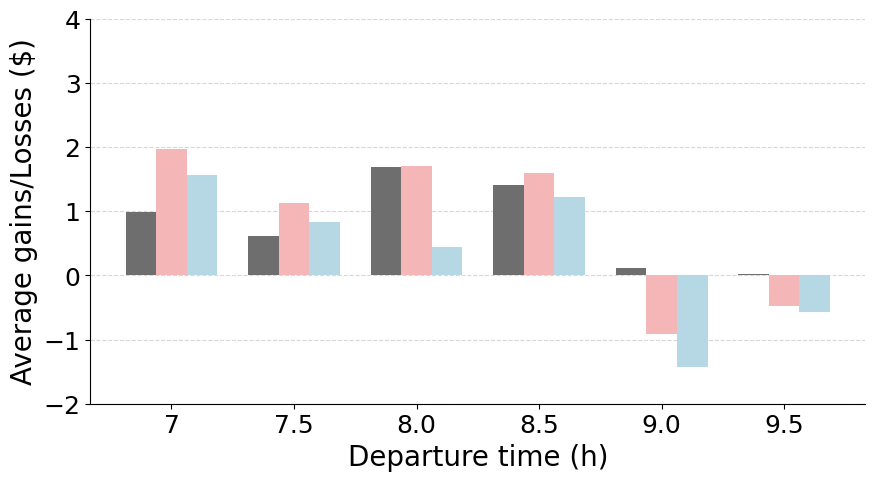}
    \caption{Gains and losses of (C) || (A)}
    \label{fig:gains2}
  \end{subfigure}
  \begin{subfigure}{0.68\textwidth}
    \includegraphics[width=\linewidth]{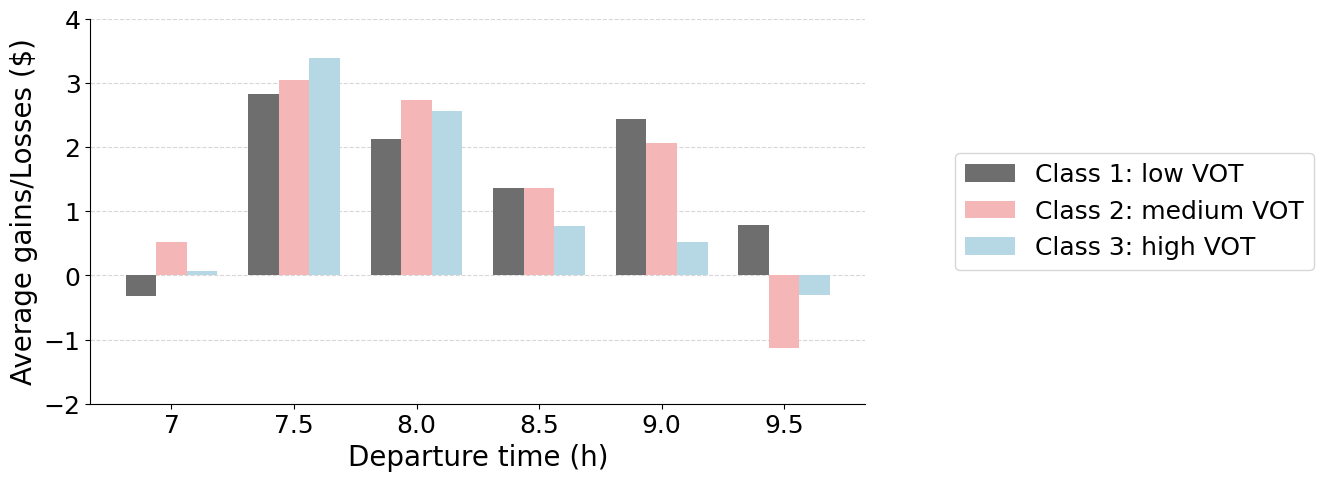}
    \caption{Gains and losses of (D) || (A)}
    \label{fig:gains3}
  \end{subfigure}
  \caption{Distribution of average commuters' gains and losses across departure times and user classes for : (1) \textit{scenario B - Incen. low VOT}, (2) \textit{scenario C - Incen. Rest. SMSs+PT}, and (3) \textit{scenario D - O-D Dyn. Incen.} relative to scenario A}
  \label{fig:gains}
\end{figure}

This section assesses equity by examining changes in travel costs. In this section, we use the double-bar (||) notation to express this relativity concept (e.g., (B) || (A) refers to the gains in scenario B relative to scenario A). Figure \ref{fig:gains} shows the average gains or losses for each user class and across departure times. Positive values represent gains in the travel costs generated by the policy, while negative values indicate losses.

\renewcommand{\thesubfigure}{\arabic{subfigure}}

In scenario B, low VOT users have the largest gains, particularly between 8:00 and 8:30, when congestion is at its highest. However, users with medium and high VOT experience only modest gains, or in some cases, even losses. This indicates that, while the policy is effective for its target group, the benefits are not evenly distributed. In scenario C, the distribution of gains is more balanced across user classes, with all groups obtaining moderate improvements during the peak period. Some losses appear after 8:30, reflecting the delayed recovery presented in the congestion analysis. Scenario D generates the highest and most consistent gains. All user groups highly benefit during the peak period, and losses are smaller than in other cases. Gains are also relatively homogeneous across user groups, suggesting that the RL agent’s strategy improved overall system efficiency without excessively favoring any single class.

\begin{figure}[h!]
  \centering
  \includegraphics[width=\linewidth]{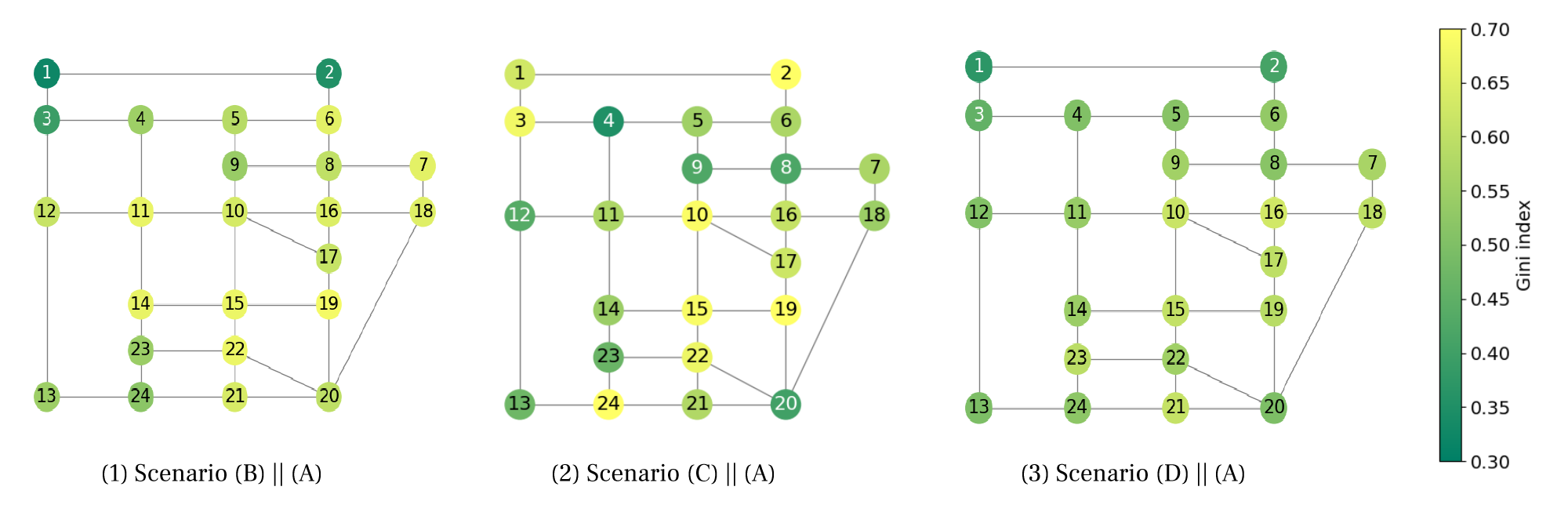}
    \caption{Spatial distribution of the Gini index across network nodes for the three incentivization scenarios. The Gini indices are measured for users originating from each node, where lower values of Gini (greener colors) signify better equity.}
    \label{fig:gini1}
\end{figure}

Moreover, equity outcomes were assessed using class-specific and global Gini indices, presented in Table \ref{tab:equity}, as well as their spatial distribution across the network (Figure \ref{fig:gini1}). These indicators show how fairly the benefits of each policy are shared across user groups and geographic areas. {The class-wise Gini coefficient presented in  Table \ref{tab:equity} measures the inequality of benefits within each class, capturing how evenly the gains are distributed among users with similar characteristics. This indicator is used to assess the internal fairness of each group.}

\begin{table}[h!]
\centering
\begin{tabular}{lccc}
\hline
& \textbf{(B) || (A) } & \textbf{(C) || (A)} & \textbf{(D) || (A)} \\
\hline
Gini index for class 1  &\textbf{ 0.59} & 0.64 & 0.60  \\
Gini index for class 2 & 0.64  & 0.60 & 0.53  \\
Gini index for class 3  & 0.61 & 0.56 & 0.55  \\
\hline
\hline
Global Gini index  &0.61 & 0.63 & \textbf{0.57}  \\
\hline
\end{tabular}
\caption{Evaluation of equity with the Gini index. (B): Incen. low VOT, (C): Incen. Rest. SMSs+PT, and (D): O-D Dyn. Incen.}
\label{tab:equity}
\end{table}

Scenario B achieves the best equity outcome for class 1, consistent with the policy design. However, equity for classes 2 and 3 remains higher, indicating that the benefits are not widely shared among all commuters. This suggests that class-targeted subsidies can increase benefits but may reinforce disparities if not accompanied by broader measures. Scenario C distributes gains more evenly between classes 2 and 3, whose Gini indices drop to 0.60 and 0.56, respectively. However, class 1 equity worsens, and the global Gini increases to 0.63, higher than scenario B. This pattern can be attributed to the design of the scenario, where incentives are provided only when there is no direct PT connection. As a result, commuters departing from the same origin but traveling to different destinations experience unequal benefits. Since low-VOT users are the most sensitive to subsidies, they are more affected by these spatial inequities. Figure \ref{fig:gini1}.(2) also reflects this aspect spatially, where few nodes show relatively higher inequality, such as nodes 2, 3, 10 and 24. 

Finally, scenario D delivers the best overall equity outcomes. Particularly, the Gini indices for classes 2 and 3 are reduced,  while the Gini index for class 1 slightly increases compared to scenario B. However, the global Gini reaches its lowest value. In this case, the benefits are distributed more evenly, avoiding the concentration of gains in a single class. Furthermore, the RL agent’s ability to vary subsidies by origin and destination introduces additional flexibility, allowing it to target areas where PT is less competitive (e.g., nodes 12 and 22) or where congestion is highest (e.g., nodes 10 and 15). Spatially, as shown by Figure \ref{fig:gini1}.(3) the results translate to a more homogeneous distribution of equity, with a relatively low Gini index across all nodes of the network.

\renewcommand{\thesubfigure}{\alph{subfigure}}

In previous experiments, all policies were evaluated under the assumption of static pricing for ridesharing. This assumption allowed us to isolate the effects of public incentives on the system performance. However, in practice, RS providers are profit-maximizing actors who continuously adjust their fares to match demand conditions, aiming to attract more passengers and increase revenues. As a result, pricing is inherently dynamic, and its interaction with public policies can significantly alter system outcomes. Thus, representing the RS provider as a dynamic learning agent is essential for evaluating the real impact of incentivization policies.

\subsection{Joint dynamic pricing and incentivization strategies}
\label{sec:joint}

In this section, we examine a joint decision-making framework in which both RS providers and public authorities deploy RL models with potentially conflicting objectives. This analysis enables us to assess the impacts of a greedy RS provider and examine how public authorities can counterbalance these effects to preserve efficiency, fairness, and environmental performance.

To present the results, we establish a set of benchmark scenarios against which the performance metrics can be compared. The scenarios considered in this section are defined as follows:

\begin{itemize}
    \item \textbf{Static pricing (SP), no incentivization}: the RS service provider does not adjust the trip fares. Thus, the dynamic price factors $\lambda^{(i,j)}$ are all set to 1. Similarly, public authorities do not offer any incentive on PT (i.e., $\delta^{(i,j)}=0 \quad \forall i,j \in E$). This represents the baseline scenario previously described (scenario A).
    \item \textbf{Dynamic pricing (DP), no incentivization}: the RS service provider uses RL to dynamically adjust the trip fares $\lambda^{(i,j)}$. However, public authorities do not offer any incentive on PT (i.e., $\delta^{(i,j)}=0 \quad \forall i,j \in E$).
    \item \textbf{Static pricing with incentivization (SP-I-TGC)}: the RS service provider does not adjust the trip fares, while public authorities use RL to offer spatio-temporal incentives on PT to improve the total generalized cost. This represents the scenario (D) described in the previous section.
    \item \textbf{Dynamic pricing with incentivization (DP-I-TGC)}: the RS provider uses RL to dynamically adjust the trip fares. Similarly, public authorities use RL to offer dynamic incentives to improve the total generalized cost.
    \item \textbf{Dynamic pricing with incentivization for emissions (DP-I-SER)}: the RS provider uses RL to dynamically adjust the trip fares. Similarly, public authorities use RL to offer spatio-temporal incentives on PT to improve the system emission rate.
    \item \textbf{Dynamic pricing with incentivization for equity (DP-I-G)}: the RS provider uses RL to dynamically adjust the trip fares. Similarly, public authorities use RL to offer spatio-temporal incentives on PT to improve equity.
\end{itemize}

Details on the hyperparameters and training performance are provided in \ref{ap:training}. {We acknowledge that additional non-RL approaches, such as heuristic or optimization-based formulations, could further strengthen the comparison. However, the problem considered here is inherently dynamic, with strong feedback loops between demand, congestion, pricing, and incentives, making the design of comparable sequential decision-making baselines non-trivial. While advanced optimization-based methods could in principle be formulated, their implementation in this simulation-based multimodal setting would require solving high-dimensional, non-convex problems at each decision step. The objective of this study is therefore not to demonstrate superiority over a specific deterministic benchmark, but to evaluate the potential of dynamic, spatio-temporal policies to capture nonlinear interactions and improve system performance. Incorporating richer baseline comparisons remains an important direction for future work.}

In what follows, we analyze the impacts on traffic congestion, as it represents the most critical aspect influencing all other performance metrics, and present a comparative evaluation of the scenarios across the different indicators.

\begin{figure}[h!]
  \centering
  \begin{subfigure}{0.38\textwidth}
    \includegraphics[width=\linewidth]{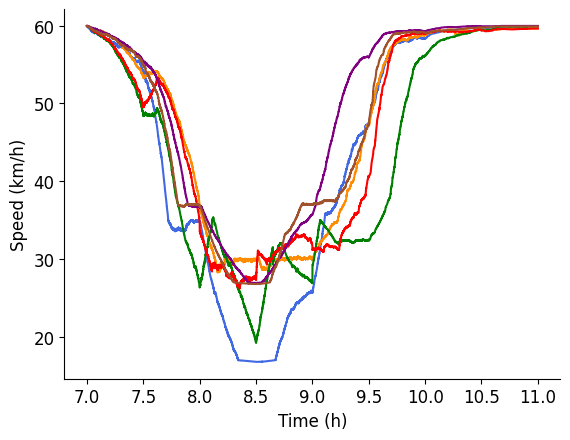}
    \caption{}
    \label{fig:benchmark1}
  \end{subfigure}
  \begin{subfigure}{0.47\textwidth}
    \includegraphics[width=\linewidth]{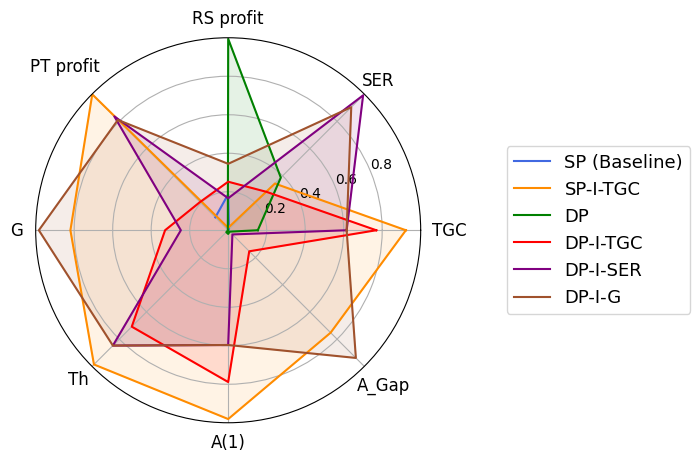}
    \caption{}
    \label{fig:benchmark2}
  \end{subfigure}
  \caption{(a): Temporal evolution of average network speed. (b): Comparative analysis of the performance metrics for the simulated scenarios. \\
  {Outward movement on the radar chart indicates improvement for all metrics, including those usually minimized}}
  \label{fig:benchmark}
\end{figure}

Figure \ref{fig:benchmark1} illustrates the evolution of average network speed, which serves as a proxy for congestion levels. The DP case demonstrates a clear improvement compared to the static baseline, as the RL agent dynamically adjusts fares to regulate ridesharing demand in response to network conditions. However, its performance remains below that of the SP-I-TGC scenario, which serves as an idealized benchmark from a system perspective. The DP curve also exhibits noticeable fluctuations, reflecting the inherent instability of a purely profit-driven pricing policy. As the RL agent seeks to maximize revenue, it periodically lowers fares to attract more demand, which can temporarily alleviate congestion by redistributing trips but also disrupt the spatial balance of vehicle supply. The resulting mismatch between supply and demand leads to an increase in unsatisfied requests, leading the agent to raise prices again. This reactive behavior creates oscillations in both demand patterns and network congestion levels. In contrast, the DP-I-TGC scenario achieves a more balanced and stable performance. By coupling dynamic ridesharing pricing with public transport incentivization, the RL agents learn to manage the trade-off between profit maximization and system-wide efficiency. The public transport incentives help absorb excess demand when ridesharing prices rise, while adaptive RS pricing offers an attractive option for relevant commuters. This coordination results in smoother traffic conditions and a faster recovery toward free-flow speeds after the peak period, compared to the DP scenario. Scenarios DP-I-SER and DP-I-G demonstrate globally a similar pattern to DP-I-TGC scenario.


Moreover, Figure \ref{fig:benchmark2} and Table \ref{tab:benchmark} present a comparative evaluation of previously described scenarios across the different performance metrics, providing a complete view of how each strategy influences the system efficiency, equity, environmental outcomes, and financial performance. Particularly, we analyze the multidimensional trade-offs in terms of TGC, profit of mobility providers, emissions, and equity. We evaluate equity across four metrics: Gini index, Theil coefficient, Atkinson index with $\epsilon=1$, and accessibility gap (see \ref{ap:metrics}).

For visual consistency of Figure \ref{fig:benchmark2}, each metric was normalized to its minimum and maximum values, resulting in a unit-free chart. This allows all metrics to be displayed on the same scale, making comparisons more intuitive. In addition, the normalization is performed so that higher values always represent better performance, meaning that even for metrics to be minimized (SER, TGC, G, Th, A(1), and A\_Gap), the model achieving the lowest value appears closest to the outer edge of the chart.

\begin{table}[h!]
\centering
\small
\begin{tabular}{lcccccccc}
\hline
& \textbf{TGC ($\times 1^5$ USD)} & \textbf{SER (Kg.h)} &  \textbf{RS profit (USD)} & \textbf{PT profit (USD)} & \textbf{G} & \textbf{Th} & \textbf{A(1)} & \textbf{A\_Gap}\\
\hline
SP (Baseline)  & 3.0 & 1198 & 4023 & 5658 & - & - & - & -  \\
SP-I-TGC & \textbf{2.4} & 1082 & 3899 & \textbf{10991} & 0.57 & \textbf{0.29} & \textbf{0.29} & -0.63  \\
DP & 2.9 & 1067 & \textbf{4596} & 5115 & 0.67 & 0.36 & 0.34 & -0.27 \\
DP-I-TGC & 2.5& 1103 & 4069 & 6322 & 0.63 & 0.31 & 0.30 & -0.34 \\
DP-I-SER & 2.6 & \textbf{859} & 4008 & 10036 & 0.64 & 0.30 & 0.31 & -0.28 \\
DP-I-G & 2.6 & 889 & 4135 & 9882 & \textbf{0.55} & 0.30 & 0.31 & \textbf{-0.72}\\

\hline
\end{tabular}
\caption{Comparative analysis of the performance metrics for the simulated scenarios.}
\label{tab:benchmark}
\end{table}

The RS profit is defined as the total fare revenues from RS trips minus the corresponding vehicle operating costs. This measure corresponds directly to the reward function of the RS provider’s RL agent, as formulated in Equation \ref{eq:RSreward}. The PT profit is computed as the total fare revenues from PT trips. Operating costs for PT are not included, based on the simplifying assumption that PT services operate independently of demand levels. Thus, including them would not affect the comparative analysis.


The results reveal that DP policy provides an improvement of 14\% on the RS profit, as a direct consequence of the provider’s revenue-maximizing behavior. Moreover, it also improves the emissions and TGC by enabling more sharing in the network compared to the baseline scenario. However, these gains are achieved at the expense of equity and public transport attractiveness. This reflects the profit-maximizing behavior of the provider, which naturally prioritizes its own revenues over societal outcomes. 

With incentivization schemes, the system dynamics shift according to the chosen optimization objective. Targeting TGC achieves the best performance on that specific metric, but this focused optimization does not automatically improve all other metrics and comes with trade-offs in dimensions that are not directly targeted. Similarly, an authority optimizing emissions drives the system toward better environmental performance and higher public transport use, but this reduces ridesharing profitability since incentives are allocated mainly to shift demand toward PT.

When considering the joint dynamic strategy, optimizing for equity produces the {best trade-off across the evaluated performance metrics in this test case.} The equity-oriented agent learns when and by how much to provide incentives so that PT becomes attractive in locations where it is most needed. Thus, PT profit increases by 74\% compared to the baseline, while equity improves, and RS services still capture a significant share of demand and maintain profitability. Furthermore, by explicitly minimizing disparities among commuters, this strategy indirectly smooths extreme losses and gains across the system, which in turn contributes to about a 13\% reduction in TGC.

In the previously described experiments, we explicitly distinguish PT profit from the profit of public authorities. While fare revenues collected from PT users are part of public income, they are treated here as PT profit. In contrast, the incentives provided by public authorities are not deducted from this amount. In the following section, we analyze the financial implications of allocating these incentives to public authorities. 

\subsection{Economic impact and sensitivity analysis}

The distinction of PT profit from the net economic outcome of public authorities reflects the common practice in which authorities finance such incentivization measures through external funding mechanisms, such as taxes or dedicated public budgets, rather than from operational fare revenues. In this context, the role of the authority is to allocate a specific budget to enhance transport system performance and societal welfare, rather than to maximize direct financial returns \citep{small_economics_2024}.

Under this distinction, it becomes particularly important to analyze the profitability of each policy. Thus, we provide in this section a comparison of system-wide monetary outcomes. Moreover, to understand how these outcomes respond to underlying modeling assumptions and behavioral heterogeneity, we also conduct a brief sensitivity analysis. 

We present in Table \ref{tab:moneyGains} an analysis of the expenses and benefits of public authorities, including incentive costs and their corresponding gains in system efficiency and emissions, as described by equation (\ref{eq:gains}) for a scenario $S$.

{It should be noted that emissions are expressed in monetary terms as part of the evaluation framework to facilitate comparison across different performance dimensions. This transformation provides a common scale to interpret trade-offs between efficiency, environmental performance, and other objectives. Moreover, the total cost of incentivization is determined endogenously by the policy, as it depends on both the level of incentives and the resulting demand response. As such, the different scenarios reflect varying levels of intervention intensity, and the analysis provides insights into the cost-effectiveness of each policy in terms of resource utilization.}

We focus on the three previously presented joint pricing and incentivization scenarios to provide the most realistic system-level perspective. For each incentivization strategy, we evaluate the trade-off between the gains, measured in terms of reduced emissions and increased PT profit compared to the baseline scenario, and the losses associated with the subsidies provided. We use the following formula to convert the SER to monetary values.
\begin{equation}
\label{eq:gains}
    \text{gains}_S = (\text{PT profit}_S - \text{PT profit}_{SP}) - \text{Incentives}_S+ (\text{EC}_S - \text{EC}_{SP}) 
\end{equation}

where \textit{S} is an incentivization scenario, \textit{SP} is the baseline scenario. We recall that PT profit is the revenue from the base fare of PT, as previously described. For each scenario, we compute:

\begin{align}
    & \text{Incentives} = \sum_{i,j \in E} \delta^{(i,j)} \cdot q_m^{(i,j)} & m \in [bus, M] \\
    & \text{EC} = \frac{EP}{1000} \cdot\sum_{t} \sum_{a \in A} 0.2038 \cdot tt_{a} \exp({\frac{0.7962 \cdot l_{a}}{tt_{a}}}) \cdot x_a \cdot \Delta T_t \label{eq:EC}
\end{align}

where in equation \ref{eq:EC}, $\Delta T_t$ represents the length of time step $t$. $EP$ is the emission price (in USD per kilogram of pollutant). In this study, we assume $EP = 0.026$ USD/Kg \citep{akerboom_meeting_2020}.

All incentivization strategies result in a negative net economic outcome for public authorities, meaning that the cost of the incentives provided exceeds the monetized gains (from emission reductions and increased PT profit) across all cases. The \textit{DP-I-SER} exhibits the lowest net financial cost, making it the least expensive strategy from the perspective of public authorities, while \textit{DP-I-G} presents the highest net cost, indicating that prioritizing equity requires the largest investment. This outcome is also partly due to the fact that equity improvements are socially valuable but cannot be directly translated into monetary terms, and therefore are not captured in the financial balance.

Although the strategies imply a financial cost for public authorities, the results suggest that such policies may still be attractive in contexts where authorities centrally manage or directly collaborate with SMSs providers, as part of the benefits could then be retained. The framework developed in this study can be used to test different incentive designs before implementation, helping decision-makers identify which strategy offers the best compromise based on their objectives. A promising direction for future research is to extend the model toward multi-objective learning, allowing the RL agent to self-balance profitability, efficiency, and equity. This could lead to more adaptive policy designs that reflect changing priorities over time.


\begin{table}[h]
\centering
\begin{tabular}{lccc}
\hline
 & \textbf{DP-I-TGC} & \textbf{DP-I-SER} & \textbf{DP-I-G}\\
\hline
Economic outcome (USD) for public authorities & -6103 & -5163 & -7827 \\
\hline
\end{tabular}
\caption{Net economic outcome for public authorities (in USD)}
\label{tab:moneyGains}
\end{table}

{The financial analysis also reveals an asymmetry in the economic outcomes. While public transport incentivization generates direct fiscal costs for the public authority, the ridesharing operator experiences profit gains, particularly under dynamic pricing. The contrast between public financial costs and private profit growth reflects a distributional imbalance within the multimodal system. 
From a regulatory perspective, this observation points to the potential of cross-subsidy mechanisms in improving the financial viability of such policies. Several instruments could be considered. Platform taxation schemes and revenue-sharing agreements could be used to capture part of the additional profit generated by dynamic pricing, while congestion-based charges applied to ridesharing could both internalize its contribution to traffic conditions and generate public revenue. Such mechanisms would also affect the fares faced by users and may induce different behavioral responses, potentially altering demand patterns and system performance. Incorporating such mechanisms into the modeling framework constitutes a promising direction for future research.}

To complement this economic evaluation, we conduct two complementary sensitivity analyses to assess the robustness of the proposed framework. The first evaluates how the pricing model responds to variations in the spatial distribution of ridesharing vehicles, examining whether changes in fleet allocation affect the resulting system performance. The second focuses on the equity concept by testing the sensitivity of the incentivization model to the distribution of users’ VOT. The results, presented in \ref{ap:SA}, reveal that the dynamic pricing model is more sensitive to the spatial distribution of the fleet rather than to its total size. Consequently, achieving a more homogeneous fleet distribution leads to better outcomes, whereas increasing the fleet size provides only limited additional benefits. Similarly, the incentivization model shows a notable sensitivity to the VOT heterogeneity. As a result, the spatial distribution of the population emerges as a key parameter influencing the outcomes of our RL-based approach. Since this distribution is embedded in the learning process, any significant change would require retraining the RL model to maintain comparable performance. 

Taken together, the presented results show that public authorities can realign system outcomes toward broader societal goals through incentivization, countering the negative effects of a profit-driven SMS provider. For the considered test case, an equity-oriented policy appears particularly effective at producing the best trade-off between equity, efficiency, and environmental performance, despite its relatively low monetary return. Moreover, the results illustrate the ability of the proposed framework to capture dynamic interactions between stakeholders, and the resulting trade-offs in terms of system performance and profitability. 

{The reported performance improvements are obtained under a specific network configuration, demand profile, and matching mechanism, and the corresponding results should therefore be interpreted within this context. Changes in network structure or scale may affect congestion patterns, availability of travel options, and the spatial distribution of demand, which in turn could influence the effectiveness of pricing and incentivization policies. Similarly, evolving demand patterns may require retraining or adaptation of the learned policies to maintain performance. In addition, the matching algorithm plays a critical role in determining service efficiency and vehicle utilization. More advanced matching and repositioning strategies could alter system dynamics and potentially amplify or attenuate the observed effects. While the proposed framework is flexible and can be applied to different settings, assessing its robustness across diverse network configurations and operational conditions constitutes an important direction for future work.}

\section{Conclusion}
\label{conclusion}

This paper proposed a multi-agent deep reinforcement learning framework to coordinate dynamic pricing strategies of ridesharing providers with spatio-temporal incentivization policies introduced by public authorities, to optimize the traveler's costs, vehicular emissions, and spatial equity in a multimodal transportation network. By modeling the interactions of diverse stakeholders within a macroscopic simulation environment, the study addresses a significant gap in existing literature, where pricing and incentivization decisions are often optimized independently or under static assumptions. The proposed framework, therefore, offers a novel approach to analyze complex dynamics while explicitly accounting for system feedback, modal interactions, user heterogeneity, intermodality, and the conflicting objectives of private and public actors. It also offers strong potential as a decision-support tool for the design of future mobility policies that promote sustainability and equity. 

{The numerical experiments highlight the limitations of static, rule-based incentivization schemes and the importance of accounting for modal interactions. For instance, targeted incentivization of low-VOT users is observed to induce a shift from public transport to ridesharing, resulting in a slight increase in emissions despite a reduction in private vehicle use. While this effect is specific to the considered setting, it illustrates how SMSs policies can generate unintended impacts when designed in isolation. This observation reinforces the need for coordinated, system-wide approaches.} In this context, the proposed dynamic incentivization strategy, jointly optimized with dynamic pricing, enables public authorities to achieve desirable trade-offs among efficiency, equity, profitability, and environmental goals. Among the examined strategies, the equity-oriented policy achieved the {best trade-off across the evaluation dimensions}, results across the evaluation dimensions, leading to a 74\% increase in public transport profit, enhancements in spatial equity, and a reduction in total generalized cost, while only reducing the profit of RS provider by 10\%, compared to a greedy setting. These findings show that promoting equity does not necessarily compromise efficiency or environmental objectives.


Future research should extend the framework to address a few key limitations. First, the framework is implemented on a medium-scale test network. Scaling the methodology to realistic metropolitan contexts would require improving computational performance. Several promising directions can be explored, such as the use of clustering techniques to partition the network, allowing direct application of the proposed model. Additionally, advanced approaches for user equilibrium assignment, such as gradient projection heuristics or genetic algorithms, could be integrated to improve the scalability and accuracy of the model. Further optimization can be achieved by enhancing the passenger-driver matching process and refining the simulation framework through more efficient data structures and parallel computing. {Furthermore, the current model assumes full compliance from ridesharing drivers and does not incorporate vehicle repositioning strategies, which are essential to operational realism. Future extensions could relax this assumption by incorporating strategic driver behavior, such as selective acceptance of trips. Potential extensions also include combining pricing and decentralized repositioning strategies within a unified reinforcement-learning model or applying optimization-based repositioning techniques in conjunction with RL.} {In addition, extending the analysis to include comparisons with non-RL baselines, such as optimization-based approaches, would provide further insights into the performance of dynamic learning-based strategies.} {Future work could also extend the framework to incorporate other equity objectives, such as social equity objectives.} {Finally, minor fluctuations observed during training suggest opportunities to improve learning stability through enhanced reward and state design, more advanced RL architectures, or alternative learning and optimization paradigms.}


\section*{Authors contribution statement}
\noindent \textbf{Khadidja Kadem}: Conceptualization, Methodology, Software, Investigation, Formal analysis, Writing - original draft, Writing - review \& editing. \textbf{Mostafa Ameli}: Conceptualization, Methodology, Formal analysis, Writing - review \& editing, Project administration, Supervision. \textbf{Carlos Lima Azevedo}: Conceptualization, Methodology, Formal analysis, Writing - review \& editing, Project administration, Supervision. \textbf{Mahdi Zargayouna}: Conceptualization, Methodology, Formal analysis, Writing - review \& editing, Project administration, Supervision. \textbf{Latifa Oukhellou}: Conceptualization, Methodology, Formal analysis, Writing - review \& editing, Project administration, Supervision.

\section{Acknowledgments}

This work (M. Ameli) was supported by the French ANR research project SMART-ROUTE under Grant ANR-24-CE22-7264.

\section*{Conflicts of interest}
None.


\setcounter{table}{0}
\setcounter{figure}{0}

\appendix

\section{List of important notations}
\label{ap:notation}
\begin{table}[H]\centering
\caption{\textbf{List of important notations}}
{\footnotesize
\begin{tabular}{p{1.2cm}p{14cm}}\hline
$E$ & Set of network nodes. \\
$A$ & Set of network links.\\
$\Psi$ & Set of all travel modes; $\Psi$ = \{PV, bus, M, CP, CD, RS, W, B, $I_s(m1,m2)$\}. \\
$K$ & Set of user classes. \\
$P_{ij}$ & Set of paths for OD pair $(i,j)$. \\
$I$ & Set of incentives offered by the public authorities. \\
$\alpha_k$ & Value of time of users' class $k$. \\
$\zeta_k$ & Proportion of car owners for class $k$. \\
$\gamma$ & Monetary cost per unit of distance. \\
$L_p$ & Length of path $p$. \\
$F_{p,m}^{(i,j)}$ & Trip fare of mode $m$ on path $p$ between OD pair $(i,j)$. \\
$b_m$ & Fixed base fare for mode $m$. \\
$Q_k^i$ & Proportion of users of class $k$ departing from node $i$. \\
$U(S_1, S_2)_k^i$ & Average benefits from a policy or scenario $S_1$ relative to a baseline scenario $S_2$ for users of class $k$ departing from node $i$. \\
$EP$ & Emission price. In this study, we assume $EP = 0.026$ USD/Kg. \\ 
\\
\textbf{\underline{Indices} }& \\
$\tau$ & Index of time (discretized) \\
$k$ & Index of users class, $k \in K$. \\
$i,j$ & Index of node, $i,j \in E$. \\
$a$ & Index of link (edge), $a \in A$. \\
$m$ & Index of mode, $m \in \Psi$. \\
$p$ & Index of path, $p \in P_{ij} $. \\
\\

\textbf{\underline{Variables}} &  \\
$f_{p,m,k}^{(i,j)}$ & Traffic flow of mode $m$ on path $p$ for users of class $k$ between OD pair $(i,j)$, (Integer variable). \\
$\lambda^{(i,j)} (\tau)$ & Ridesharing price factor defined by the service provider at time-step $\tau$ for OD pair $(i,j)$. \\ 
$\delta^{(i,j)} (\tau)$ & Incentive offered by the authorities to public transport users at time-step $\tau$ for OD pair $(i,j)$. \\
$c_{p,m,k}^{(i,j)}$ & Generalized cost of path $p$ with travel mode $m$ between OD pair $(i, j)$ for user class $k$. \\ 
$TT_{p}$ & Travel time on path \textit{p}. \\
$WT_{m}$ & Waiting time for travel mode $m$. \\
$p_{ij}^\tau$ & PT demand between the OD pair $(i,j)$ at time $\tau$. \\
$d_{ij}^\tau$ & Ridesharing demand between the OD pair $(i,j)$ at time $\tau$. \\
$x_a$ & Vehicular flow of link $a$. 
\end{tabular} } 
\end{table}

\section{Boundary discontinuity analysis for dynamic incentivization}
\label{ap:BI}
To assess potential boundary effects induced by the node-based decomposition of the action space, we conduct an experimental analysis on the spatial consistency of the learned incentivization policy. In particular, we examine whether adjacent nodes receive significantly different incentive levels, which could indicate artificial discontinuities introduced by the discretization.

We quantify boundary discontinuities by measuring the difference in incentives assigned to neighboring nodes. For each pair of adjacent origin nodes $a \sim a'$, and for a given destination $b$, we compute the absolute difference in incentives:

\begin{equation}
\Delta_{\text{origin}}  (a, a';b) = \left| \delta^{(a,b) }- \delta^{(a',b) }\right| = \frac{1}{2} \left| \delta_o^{a}- \delta_o^{a'}\right|
\end{equation}

Similarly, for each pair of adjacent destination nodes $b \sim b'$, and for a given origin $a$, we compute:

\begin{equation}
\Delta_{\text{destination}}  (a;b, b') = \left| \delta^{(a,b) } - \delta^{(a,b') }\right| = \frac{1}{2} \left| \delta_d^{b}- \delta_d^{b'}\right|
\end{equation}

We report in Table \ref{tab:discontinuity} the statistical distribution of incentive discontinuities across adjacent nodes. At the origin level, the average discontinuity is \$1.19, while at the destination level it reaches \$1.81, with higher values also observed in the upper percentiles. This asymmetry indicates that the learned policy varies more strongly with destination than with origin, suggesting that incentives are more driven by destination attractiveness than by the departure location.

\begin{table}[h]
\centering
\caption{Summary statistics of incentive discontinuities across neighboring nodes}
\label{tab:discontinuity}
\begin{tabular}{lcc}
\hline
Statistic & $\Delta_{\text{origin}}$ & $\Delta_{\text{destination}}$ \\
\hline
Mean     & 1.1895 & 1.8079 \\
Std Dev  & 0.6946 & 1.1508 \\
Max      & 1.9000 & 3.2000 \\
\hline
\end{tabular}
\end{table}

\begin{figure}[h!]
  \centering
  \begin{subfigure}{0.47\textwidth}
    \includegraphics[width=\linewidth]{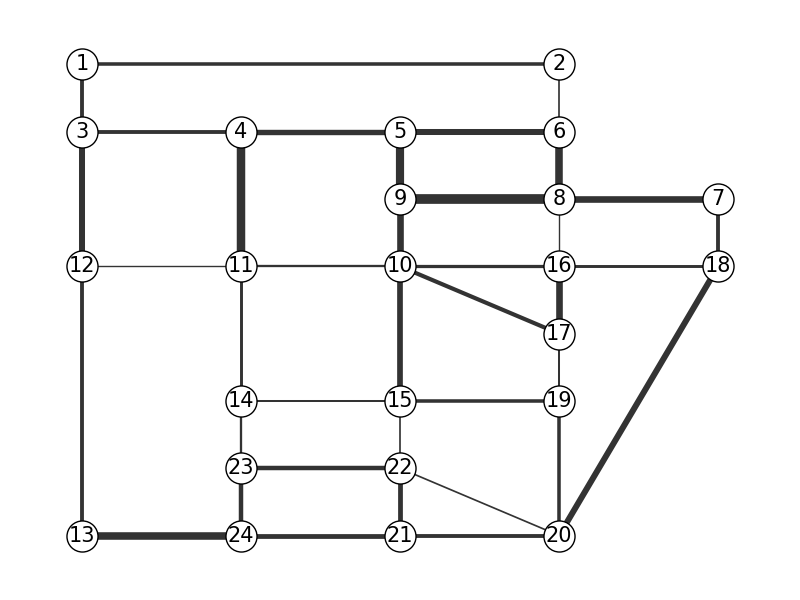}
    \caption{Origin-level discontinuity}
    \label{fig:discontinuity1}
  \end{subfigure}
  \begin{subfigure}{0.47\textwidth}
    \includegraphics[width=\linewidth]{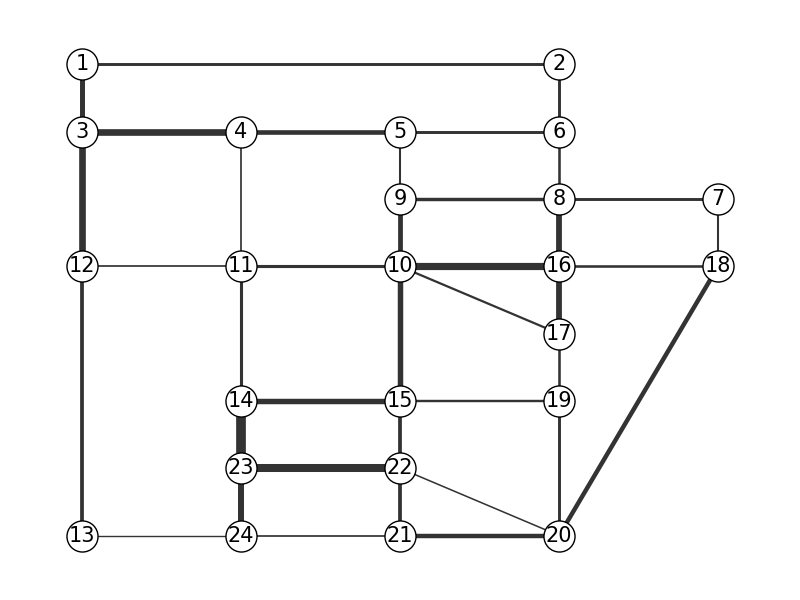}
    \caption{Destination-level discontinuity}
    \label{fig:discontinuity2}
  \end{subfigure}
  \caption{Spatial distribution of incentive discontinuities across neighboring nodes. Edge thickness is proportional to the discontinuity between two adjacent nodes}
  \label{fig:discontinuity}
\end{figure}

We also display in Figure \ref{fig:discontinuity} the spatial distribution of the observed discontinuities across the network. The results show that these discontinuities are not uniformly distributed across the network. They are primarily concentrated in the central area, particularly around node 10 (which corresponds to the highest-demand region), and at interfaces between nodes of different classes (e.g., links such as 3-12, 23-22, and 13-24). These locations correspond to transitions in demand patterns.

Importantly, such discontinuities cannot be solely attributed to the node-based decomposition. They may also reflect underlying variations in the transportation system, including differences in public transport availability and the spatial distribution of shared mobility supply. As a result, variations in incentives across neighboring nodes can emerge both from the modeling approach and from the intrinsic heterogeneity of the network captured by the RL agent. 

To assess whether these boundary effects translate into meaningful disparities, we compare them with the spatial distribution of benefits presented previously in Figure \ref{fig:gini1}.3. Although this figure was not specifically designed to evaluate boundary consistency, it indicates that adjacent nodes do not exhibit significant differences in their Gini indices. This suggests that the overall distribution of benefits remains relatively homogeneous across neighboring nodes, despite localized variations in incentive levels. Therefore, the observed discontinuities do not appear to introduce significant spatial biases in the system.

Finally, while the node-based decomposition provides a tractable way to reduce the dimensionality of the action space, this analysis highlights the existence of localized discontinuities. A promising direction for future work is to incorporate spatial regularization mechanisms or continuous representations to enforce smoother transitions between neighboring nodes while preserving responsiveness to local conditions. 

\section{Equity metrics}
\label{ap:metrics}

In this section, we present several equity metrics to evaluate how transportation benefits are distributed among different commuter classes ($k$) and geographic regions ($i$).  Throughout this section, $Q_k^i$ is the proportion of users of class $k$ departing from node $i$. $U_k^i = U(S_1,S_2)_{k}^i$ is the average benefit that class $k$ users from node $i$ receive in policy/scenario $S_1$ relative to baseline $S_2$. The mean benefit across all users is 
    \[
    \bar{U} = \frac{\sum_{i \in E} \sum_{k \in K} Q_k^i \cdot U_k^i}{\sum_{i \in E} \sum_{k \in K} Q_k^i}.
    \]
For group-based metrics, $\bar{U}_{\text{advantaged}}$ and $\bar{U}_{\text{disadvantaged}}$ are the mean benefits computed over a subset of users (e.g., high-income vs. low-income) using the same weighting by $Q_k^i$ as in $\bar{U}$.

Given these notations, we define the equity metrics as follows:

\begin{itemize}
    \item \textbf{Theil coefficient}: a decomposable metric that calculates within-group and between-group inequality, which allows for a more detailed analysis of inequity in a system \citep{gao_regulating_2024}. 
    \[
    Th = \text{Within} + \text{Between} = \sum_{i \in E} \sum_{k \in K}  \frac{Q_k^i U_k^i}{N \cdot U_k} \cdot\ln(\frac{U_k^i}{U_k} ) +  \sum_{k \in K}  \frac{Q_k U_k}{N \cdot \bar{U}} \cdot\ln(\frac{U_k}{\bar{U}} )
    \]

    $Th$ ranges from 0, representing perfect equality, to $\ln(N)$ in the case of perfect inequality.

    \item \textbf{Atkinson index}: measures the fraction of total benefits the society would sacrifice for perfect equality, given an inequality aversion $\epsilon$. It is defined as follows:
    \[
    A(\epsilon) = 1 - \frac{\left[ \frac{1}{N} \sum_{i \in E} \sum_{k \in K} \left(U_k^i\right)^{1 - \epsilon} \right]^{\frac{1}{1 - \epsilon}}}{\bar{U}},
    \]
    $\epsilon$ represents the sensitivity to fairness. In other words, a higher $\epsilon$ gives more weight to the lowest-benefit individuals. $A(\epsilon)$ ranges from 0 to 1, which represents a strong inequality.
    \item \textbf{Accessibility gap}: measures direct benefit difference between two groups.
    \[
    \text{A\_Gap} = \bar{U}_{\text{advantaged}} - \bar{U}_{\text{disadvantaged}},
    \]
    where each group’s mean is computed as in $\bar{U}$ but restricted to the nodes/classes defining the group. This metric is not bounded, and depends on the scale of $\bar{U}$. A negative value reflects that low-VOT commuters gain more from the policy, thereby supporting social equity by prioritizing benefits toward disadvantaged populations.
\end{itemize}

It is worth noting that a wide range of equity metrics exists in the literature. Carefully designing such metrics remain an open challenge in the transportation field, particularly for multimodal networks. Some studies, such as \cite{ma_multi-dimension_2026}, propose advanced equity indices that capture accessibility gaps across different modes and are calibrated using real-world data. Our framework could benefit from such high-granularity indicators to better reflect the full complexity of benefit distributions. However, integrating these metrics would substantially increase model complexity and make the RL agent’s learning task more difficult. Therefore, we leave this aspect for future research.

\section{Passenger-vehicle matching algorithm}
\label{ap:matching}

\begin{figure}[ht]
    \centering
    \includegraphics[width=0.8\linewidth]{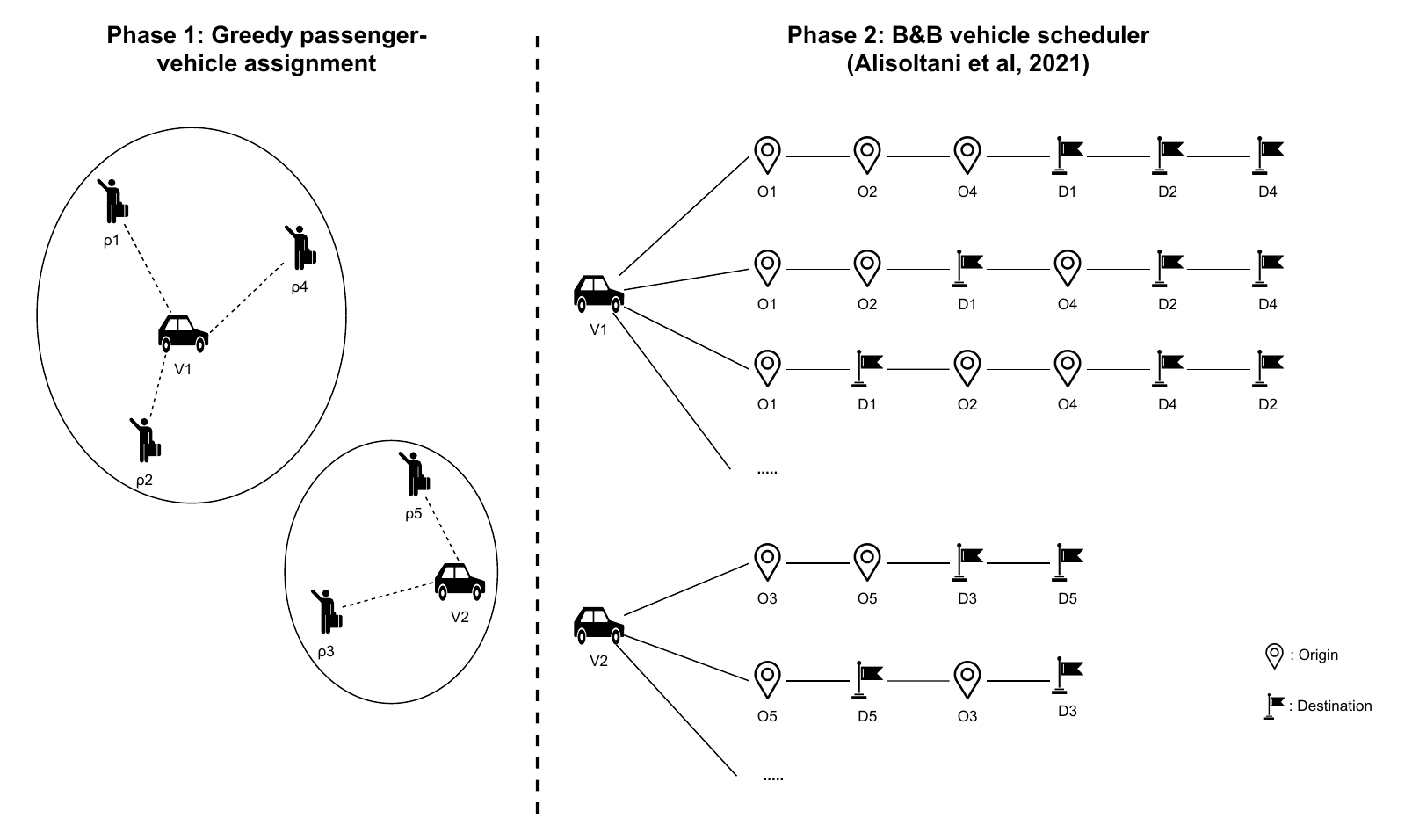}
    \caption{Passenger-driver matching procedure for carpooling and ridesharing}
    \label{fig:matching}
\end{figure}

The SMSs providers handle individual trip requests and available fleet vehicles using a matching algorithm. We adopt a two-phase approach that couples a greedy passenger-vehicle assignment with a parallelized branch-and-bound (B\&B) scheduler, inspired by \cite{alisoltani_can_2021}. Figure \ref{fig:matching} illustrates the matching algorithm.

\begin{itemize}
    \item \textbf{Greedy passenger-vehicle assignment}: Each passenger $\rho$ will submit a request with an origin node ($o_\rho$), a destination node ($d_\rho$), the earliest pickup time ($e_\rho$) and the latest arrival time ($l_\rho$). Then, each request is assigned to the vehicle whose current position (either the depot or the last scheduled stop) minimizes the travel time to $o_\rho$. This procedure is executed once for the entire batch, yielding an initial partition of passengers per vehicle $\{P_v\}_{v \in V}$ where $V$ is the set of vehicles.

    In \cite{alisoltani_can_2021}, requests are clustered based on a shareability index, and then a B\&B algorithm is executed for each group. This clustering-based approach produces accurate and globally optimized schedules. However, clusters are not independent and cannot be processed in parallel because the availability of vehicles affects the solution of each cluster. In our approach, the greedy split enables us to process vehicles independently, allowing us to evaluate their B\&B trees in parallel on separate CPU cores. This design choice is mainly motivated by the need to repeatedly execute the matching algorithm during the training of RL agents in our modeling framework. Reducing computational complexity at this stage is therefore critical to maintaining tractable training times.

    \item \textbf{Branch‑and‑Bound vehicle scheduler}: For each vehicle $v$ we solve a scheduling problem on the assigned passengers $P_v$. We use the B\&B algorithm to find the best schedule (i.e., sequence of stops) for each vehicle. The schedules are then executed using the shortest paths between stops. In the following, we provide a brief overview of the B\&B algorithm and refer the interested reader to \cite{alisoltani_can_2021} for the complete mathematical formulation.

    Each branch represents a partial stop sequence
    $<s_1,s_2,\ldots,s_q>$, where $s_i$ is either a pickup $o_\rho$ or drop‑off $d_\rho$. Branches are built incrementally by adding feasible points in terms of: 
    \begin{itemize}
        \item \textbf{Time‑window feasibility}: each passenger $\rho$ should be picked-up after its earliest departure time $e_\rho$ and dropped-off before its latest arrival time $l_\rho$.
        \item \textbf{Capacity constraint}: On‑board passenger count never exceeds vehicle capacity.
        \item \textbf{Precedence constraint}: For every request $\rho$, the destination point $d_\rho$ should appear only after the origin point $o_\rho$.
    \end{itemize}
    
    The objective function of a branch $b_v$ for vehicle $v$ is calculated as follows:
    \begin{equation}
        Obj(b_v) = \sum_{\rho \in P_v} (TT_\rho + WT_\rho) + TT_v
    \end{equation}
    where $TT_\rho$ and $WT_\rho$ are travel and waiting time of passenger $\rho$, respectively. $TT_v$ is the travel time of the vehicle. 
    
    Next, the node with the lowest lower bound is expanded; branches whose bound exceeds the current best are discarded. 
\end{itemize}

This two-phase approach balances computational performance, achieved through the greedy heuristic and parallel processing, with solution quality, via the exact enumeration of feasible stop sequences.

{It should be noted that vehicle availability is not uniformly guaranteed across space and time, since the spatial distribution evolves endogenously through the passenger-vehicle matching process and through the execution of assigned trips. At any given time, a vehicle may be idle, partially occupied (and thus still available for additional passengers), or fully occupied and therefore temporarily unavailable. After completing a service, a vehicle remains at its destination until it is assigned to a new request; if no request is assigned, it returns to the nearest depot. Consequently, localized supply shortages may emerge during peak-demand periods, generating endogenous spatio-temporal mismatches between demand and available ridesharing supply. The dynamic pricing strategy presented in Section \ref{pricing} accounts for these imbalances by adjusting fares to influence demand patterns and indirectly encourage a spatial alignment between trip requests and the current distribution of available vehicles.}

\section{Experiments configuration}
\label{ap:training}
All experiments were executed on a server equipped with an Intel Xeon E5-2680 CPU operating at 2.5 GHz. For the training, we use the Soft Actor Critic (SAC) algorithm \citep{haarnoja_soft_2018} to train the service provider agent, and the Proximal Policy Optimization (PPO) algorithm \citep{schulman_proximal_2017} for the public authorities agent. 
Table \ref{tab:params} summarizes the key parameters for our simulations.




\begin{table}[ht]
\centering
\caption{Simulation parameters setting}
\begin{tabular}{lllll}
\hline
\textbf{Parameter} & \textbf{Symbol} & \textbf{Value} \\
\hline
Time step for system evolution (min) & $\Delta T$ & 20 \\
Free-flow speed (km/h) & $V_f$ & 40 \\
Metro speed & -- & 40 \\
Walking speed & -- & 3 \\
Bike speed & -- & 10 \\
Headway of bus (min) & $1/freq_{bus}$ & 10 \\
Headway of metro (min) & $1/freq_{M}$ & 7 \\
Matching cluster size & -- & 50 \\
RS vehicle capacity & $C_{\text{RS}}$ & 2 \\
CP vehicle capacity & $C_{\text{CP}}$ & 1 \\
Bus vehicle capacity & $C_{\text{bus}}$ & 25 \\
Metro vehicle capacity & $C_{\text{M}}$ & 200 \\
SMSs base fare & $b_{m} (USD/Km) \; m \in \{ CP, RS\}$ & 0.4 \\
PT base fare & $b_{m} (USD) \; m \in \{ bus, M\}$ & 3.5 \\
Fleet size & $V$ & 1200 \\
Value of time per class & $\alpha_k \; k \in \{1,2,3\}$ & $\{5, 8, 13\}$ \\
Proportion of car owners & $\zeta$ & 70\% \\
Convergence threshold for rolling horizon timestep & - & 10\% \\
Price factors & $\lambda^{(i,j)}$ & $\in [0.2,2]$ \\
Incentives (USD) & $\delta^{(i,j)}$ & $\in \{ 0, 2.5, 5, 6\}$ \\
Mini-batch sample size & - & 80 \\
Replay buffer size & - & 100000 \\
Learning rate & - & 0.003 \\
Discount factor & - & 0.8 \\
\hline
\end{tabular} 
\label{tab:params}
\end{table}


{The convergence threshold of 10\% is associated with the internal simulation-assignment loop. This procedure relies on aggregate performance indicators, namely average network speeds and waiting times, which are iteratively updated to approximate a stable demand-supply equilibrium. In this context, the 10 \% threshold corresponds to bounded absolute variations (on the order of a few km/h for speeds and tens of seconds for waiting times), which we consider sufficiently small for practical convergence of the assignment. Tightening this threshold would lead to only marginal changes in the resulting equilibrium while significantly increasing computational time, as this procedure is executed repeatedly within each training episode. The selected value, therefore, reflects a trade-off between numerical stability and computational efficiency.}

{Due to the substantial computational burden associated with multi-agent learning in our specific context, we adopt a two-stage training procedure that separates the learning dynamics of pricing and incentivization while still accounting for their interaction. In the first stage, pricing and incentivization agents are trained independently in a stationary environment, where the behavior of the other stakeholder is fixed. This phase allows each agent to learn a stable policy under well-defined system dynamics. A single RL agent represents the ridesharing operator, which is trained for 1000 episodes to maximize profit through dynamic pricing, assuming the absence of any incentivization policy. In parallel, three separate RL agents represent the public authority, each trained independently for 1000 episodes under a static pricing scheme, with distinct optimization objectives: minimizing total generalized cost (TGC), minimizing emissions (SER), and minimizing the Gini index (G). This initial phase results in four baseline models, each converged with respect to its own objective.}

{In the second stage, the interaction between private pricing and public incentivization is explicitly introduced, since both agents are activated simultaneously and allowed to interact within the shared environment. Each incentivization agent is paired with a fixed, pre-trained pricing agent and further trained for an additional 500 episodes, allowing it to adapt its policy to the pricing behavior of the ridesharing operator. Symmetrically, the pricing agent is further trained for 500 episodes in the presence of an incentivization policy optimized for TGC. By separating individual learning from interactive learning, this approach enables each agent to adjust its strategy in response to the behavior of the other, thereby reflecting the coupled nature of pricing and incentivization in multimodal systems.}

{This sequential training approach is often adopted to mitigate the instability typically observed in multi-agent RL approaches (e.g., \cite{nekoei_dealing_2023, gerstgrasser_meta-rl_2022}), where concurrent learning induces a non-stationary environment for each agent. Pre-training each agent separately reduces oscillatory behavior and improves convergence toward a stable joint regime.}

\begin{figure}[h!]
  \centering
  \begin{subfigure}{0.45\textwidth}
    \includegraphics[width=\linewidth]{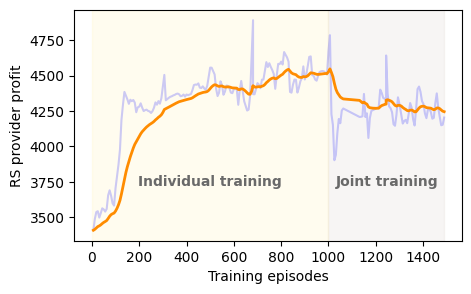}
    \caption{}
    \label{fig:train1}
  \end{subfigure}
  \begin{subfigure}{0.45\textwidth}
    \includegraphics[width=\linewidth]{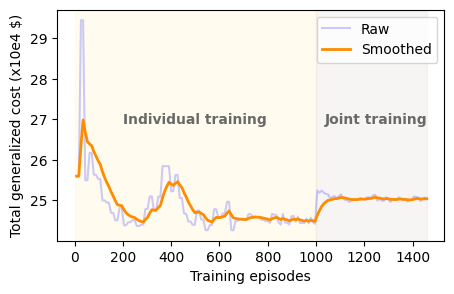}
    \caption{}
    \label{fig:train2}
  \end{subfigure}
  \begin{subfigure}{0.45\textwidth}
    \includegraphics[width=\linewidth]{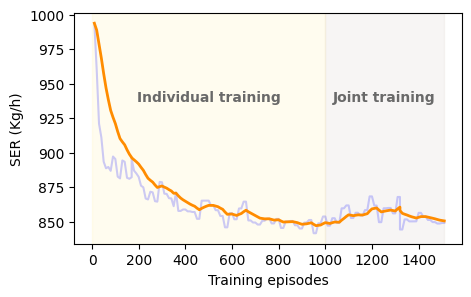}
    \caption{}
    \label{fig:train3}
  \end{subfigure}
  \begin{subfigure}{0.45\textwidth}
    \includegraphics[width=\linewidth]{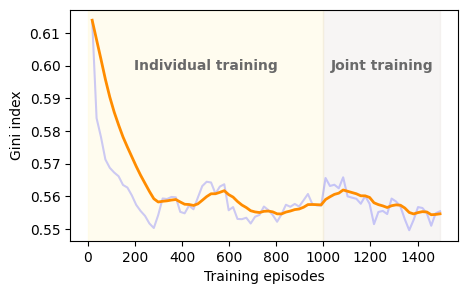}
    \caption{}
    \label{fig:train4}
  \end{subfigure}
  \caption{Convergence of the RL models. (a) Dynamic pricing model (RL agent 1) trained to maximize the ridesharing provider’s profit. (b-d) Separate RL agents (RL agent 2) representing public authorities, minimizing TGC, SER, and G, respectively}
  \label{fig:train}
\end{figure}

Figure \ref{fig:train} illustrates the convergence process of the RL models. In the figure, we present the raw outputs obtained directly from the simulations, as well as a smoothed version, calculated as a moving average over every 10 episodes, highlighting the overall trend of the learning process. Each training episode required approximately 25 minutes of computation. 

{It is important to note that some models still exhibit residual fluctuations during the joint training phase, reflecting the non-stationary state induced by the interaction between both RL agents. As a result, strict convergence is not always achieved, and training reward curves may retain some variability. In this study, all models are trained under identical configurations and for the same number of episodes to ensure comparability of results across scenarios. Within this controlled setting, the residual fluctuations are accepted as part of the learning dynamics and do not prevent consistent evaluation of relative performance. If a specific objective is of particular interest, the corresponding model can be further trained until a higher level of convergence is achieved.}

\section{Learned incentive policy: actions versus state variables}
\label{ap:inc_act}
\begin{figure}[h!]
  \centering
  \begin{subfigure}{1\textwidth}
    \includegraphics[width=0.98\linewidth]{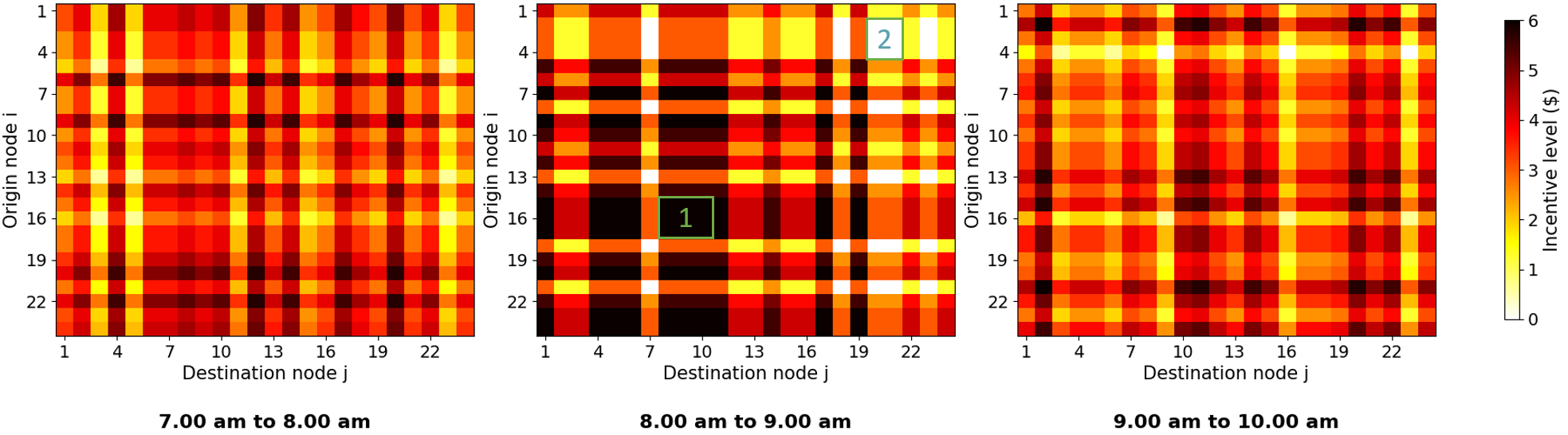}
    \caption{Spatio-temporal distribution of PT incentives}
    \label{fig:act_demand1}
  \end{subfigure}
  \begin{subfigure}{1\textwidth}
    \includegraphics[width=\linewidth]{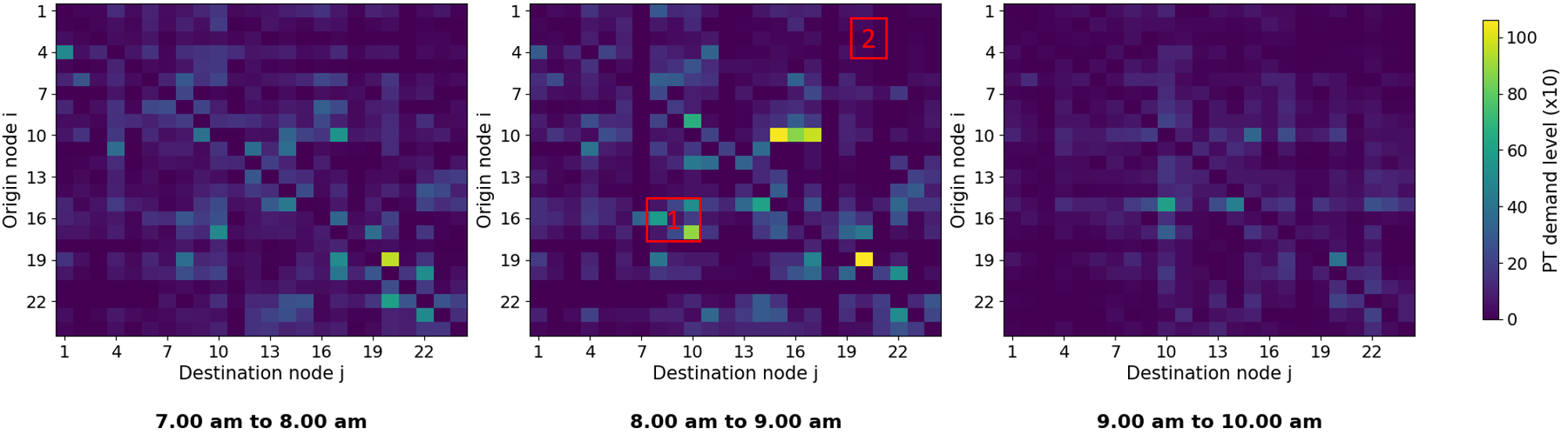}
    \caption{Spatio-temporal distribution of PT demand}
    \label{fig:act_demand2}
  \end{subfigure}
  \caption{OD-level PT incentive maps (top) and the corresponding PT demand maps (bottom) across three successive time windows of the morning peak period.}
  \label{fig:act_demand}
\end{figure}

To further interpret the behavior of the public authority RL agent, we analyze the spatio-temporal distribution of the learned incentive actions in relation to the observed PT demand, which constitutes the primary state variable of this agent. Figure \ref{fig:act_demand} presents the OD-level incentive maps (top) and the corresponding PT demand maps (bottom) across three successive time windows of the morning peak period.

Several patterns emerge from this comparison. First, the RL agent broadly increases incentive levels during the 8:00-9:00 peak window, in response to the overall rise in PT demand across the network. This is particularly evident for the OD pairs highlighted in zone 1 of Figure \ref{fig:act_demand}, where both PT demand and incentive levels reach high values. Outside this pean window (i.e., 7:00-8:00 and 9:00-10:00), we observe a more spatially uniform incentive structure as PT demand declines. This adaptive behavior reflects the agent's capacity to modulate the intensity and spatial scope of its policy in response to evolving network conditions.

Second, certain destination nodes consistently attract high incentive levels across time periods. In particular, node 10 stands out as a persistent high-incentive destination throughout the simulation horizon. This node corresponds to a highly congested area of the network, and the agent appears to have learned that directing PT users toward this destination through incentives helps relieve road congestion in its vicinity. Similarly, node 4 receives notably high incentives during the peak period, which is consistent with its role as a designated intermodal transfer point in the network.

Third, the agent assigns low or near-zero incentives to OD pairs with negligible PT demand, as illustrated by zone 2 in the figure. This behavior indicates that the agent has learned to avoid allocating budget to OD pairs where PT is already sufficiently attractive or where demand is low, thereby concentrating resources where incentives are most likely to produce a behavioral shift. 

However, it should be noted that the relationship between actions and state variables is not straightforward in our framework, due to the multi-agent and highly coupled dynamics of the system. Both RL agents act simultaneously on a shared environment, and their actions jointly influence multiple endogenous components, including mode choices, matching outcomes, and congestion levels. This coupling makes it difficult to directly attribute specific incentive decisions to individual state variables in a causal manner. A more explicit analysis of action-state relationships would require dedicated and controlled experiments, which we consider an interesting direction for future research.

\section{Sensitivity analysis}
\label{ap:SA}

We present in this section two complementary sensitivity analyses to assess the robustness of the proposed framework to variations in the spatial distribution of ridesharing vehicles, as well as variations in the spatial distribution of users’ VOT in the network. 

\subsection{RS fleet distribution}
In this section, we examine how the ridesharing pricing model responds to different fleet distribution configurations. The objective is to evaluate the model's sensitivity to both the spatial distribution and size of the RS fleet. Three scenarios were tested:

\begin{itemize}
    \item \textbf{Uniform}: service vehicles are evenly distributed across all nodes.
    \item \textbf{Center}: service vehicles are concentrated in the central nodes of the network, representing an unbalanced spatial distribution. The considered nodes are 8, 9, 10,11, and 15, where the demand is relatively the highest.
    \item \textbf{Uniform-Large fleet}: vehicles are uniformly distributed, but the overall fleet size is increased by 10 times.
\end{itemize}

\begin{figure} [h]
    \centering
    \includegraphics[width=0.75\linewidth]{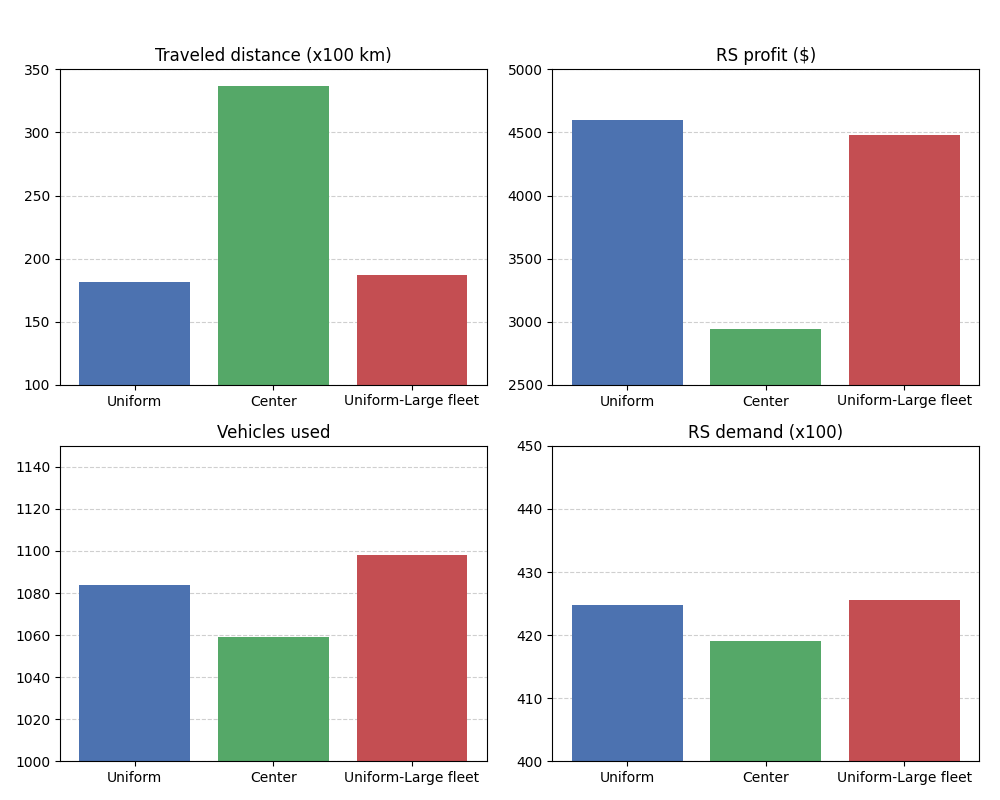}
    \caption{Sensitivity analysis of the pricing model to the RS fleet distribution. Evaluation in terms of total distance traveled by the service vehicles, provider's profit, number of service vehicles used, and RS demand}
    \label{fig:SA_fleet}
\end{figure}

Each configuration was evaluated based on four performance metrics: total distance traveled by the service vehicles, RS profit, number of vehicles used, and RS demand. Figure \ref{fig:SA_fleet} shows the results.

The traveled distance increases substantially in the \textit{center} scenario, suggesting that concentrating vehicles in central zones leads to longer trips, mostly performed empty, to serve peripheral demand. The RS profit is higher under the \textit{uniform} and \textit{uniform-large fleet} scenarios, while it drops significantly when vehicles are concentrated in the center. This implies that a balanced spatial distribution allows the model to serve demand more efficiently and profitably. The vehicles used remain relatively stable across all scenarios, with only a slight increase when the fleet size is larger. Similarly, the RS demand (i.e., the number of served requests) shows small variations. These findings suggest that the pricing model is more sensitive to the spatial distribution of the fleet rather than to its total size. Moreover, the results are also driven by the assumption that the ridesharing provider operates with a fleet of drivers with a centralized management strategy. Under this assumption, drivers always follow the system’s dispatch decisions and are directed toward the most promising locations without any delay. Consequently, the fleet capacity has little effect on the overall level of demand, since vehicle availability is effectively guaranteed. The potential impact of non-compliant or strategically behaving drivers is left for future research.

\subsection{VOT distribution: equity analysis}
In this section, we evaluate how the incentivization model responds to different VOT configurations. The objective is to evaluate the model's sensitivity to equity in terms of socioeconomic disparities among the population. We simulate four scenarios:

\begin{itemize}
    \item \textbf{Normal}: the baseline VOT distribution representing a heterogeneous mix of income groups, as depicted previously in Figure \ref{fig:SF}.
    \item \textbf{All class 1}: a homogeneous scenario where all commuters are of class 1. In other words, we consider that the whole population has low income levels.
    \item \textbf{All class 3}: another homogeneous scenario where all commuters are of class 3. We thus consider that the whole population has high income levels.
    \item \textbf{Double}: a scenario where the distribution of VOT is similar to the \textit{Normal} scenario, but every commuter’s VOT is doubled. This represents a case where commuters have a higher sensitivity to time (rather than price).
\end{itemize} 

Table \ref{tab:SA_VOT} reports the results using several equity metrics. The Gini index quantifies overall inequality in the distribution of gains, with higher values reflecting greater inequality. The Theil coefficient analyses whether disparities arise primarily within social groups (i.e., spatial equity) or between them (i.e., social equity). The Atkinson index, evaluated for different levels of inequality aversion, captures societal sensitivity to inequities. Finally, the accessibility gap represents the difference in average accessibility between the lowest and highest VOT classes.

We note that the model shows more sensitivity to VOT heterogeneity rather than its absolute scale. In other words, the differences in VOTs between commuters' classes matter more than the value level itself. This is suggested by the relatively small variations in all equity metrics between the \textit{normal} and \textit{double} scenarios, compared to the changes between \textit{all class 1} and \textit{all class 3}). Moreover, when the population is assumed to be homogeneous (scenarios \textit{all class 1} and \textit{all class 3}), the Gini index exhibits greater variation than the Theil coefficient because this latter becomes less sensitive when between-group differences are absent and only within-group variation remains. Consequently, Gini is more reactive to spatial imbalances than Theil when social heterogeneity is removed.

Additionally, although the Gini index is not explicitly designed to isolate social inequality, the observed differences across income-based scenarios (Class 1 vs Class 3) demonstrate that it still captures disparities correlated with social heterogeneity.

\begin{table}[h!]
\centering
\caption{Sensitivity analysis of the incentivization model to the VOT distribution. Equity evaluation in terms of Gini index, Theil coefficient, Atkinson index, and the accessibility gap}
\begin{tabular}{lcccccc}
\hline
\textbf{Scenario} & 
\textbf{Gini index} & 
\multicolumn{2}{c}{\textbf{Theil coefficient}} & 
\multicolumn{2}{c}{\textbf{Atkinson Index}} & 
\textbf{Accessibility gap} \\
\cline{3-4} \cline{5-6}
 &  & Within & Between & $\epsilon = 0.1$ & $\epsilon = 1$ &  \\
\hline
Normal   & 0.55 & 0.18 & 0.12 & 0.15 & 0.31 & $-0.72$ \\
All class 1  & 0.59 & 0.17 & --   & 0.14 & 0.31 & - \\
All class 3  & 0.61 & 0.16 & --   & 0.14 & 0.34 & -  \\
Double   & 0.55 & 0.20 & 0.12 & 0.15 & 0.33 & $-0.57$ \\
\hline
\end{tabular}
\label{tab:SA_VOT}
\end{table}

The Theil coefficient decomposition confirms that most inequality arises within classes rather than between them.
This indicates that even within a socioeconomic group, spatial factors (e.g., OD-based access or modal patterns) continue to be the primary source of inequity. 

For the Atkinson index, \textit{All class 3} scenario presents the highest score ($A(\epsilon=1) = 0.34$). This indicates that, even when all commuters are assumed to have high income levels, a significant proportion experiences significantly lower relative gains. This may correspond to travelers in congested zones who receive little improvement from the incentivization scheme, especially due to their high sensitivity to travel times.

The accessibility gap also supports these patterns. In the \textit{normal} and \textit{double} cases, the gap is negative, meaning disadvantaged users enjoy larger relative gains. However, the \textit{double} scenario is relatively worse in terms of accessibility. This is consistent with the policy design. Because commuters are less sensitive to price, incentives from authorities do not encourage them to use public transport. However, the commuters who were already using PT still benefit from the reduction in the trip fares.

\end{sloppypar} 

\typeout{}
\bibliography{references}
\appendix 
\end{document}